\pdfoutput=1

\documentclass[11pt]{article}

\usepackage[final]{acl}

\usepackage{times}
\usepackage{latexsym}
\usepackage{tikz}

\usepackage[T1]{fontenc}

\usepackage[utf8]{inputenc}

\usepackage{microtype}

\usepackage{inconsolata}

\usepackage{graphicx}

\usepackage{xcolor}
\usepackage{xcolor,colortbl}
\definecolor{hidden-draw}{RGB}{0,0,0}

\usepackage{tikz}
\usepackage[edges]{forest}

\usetikzlibrary{trees,positioning,shapes,shadows,arrows.meta}

\title{PlanGenLLMs: A Modern Survey of LLM Planning Capabilities}

\author{Hui Wei,$^\dagger$ Zihao Zhang,$^\ddagger$ Shenghua He,$^\S$ Tian Xia,$^\S$\\ 
\textbf{Shijia Pan,$^\dagger$ Fei Liu$^\ddagger$}\\[0.8em]
$^\dagger$University of California, Merced \, 
$^\ddagger$Emory University\, 
$^\S$PAII Inc.\\
\texttt{\{huiwei2, span24\}@ucmerced.edu}\\
\texttt{\{zihao.zhang, fei.liu\}@emory.edu}
\texttt{\quad \{shenghh2015, TianXia0209\}@gmail.com} \\ 
}

\begin{document}
\maketitle
\begin{abstract}

LLMs have immense potential for generating plans, transforming an initial world state into a desired goal state. A large body of research has explored the use of LLMs for various planning tasks, from web navigation to travel planning and database querying. However, many of these systems are tailored to specific problems, making it challenging to compare them or determine the best approach for new tasks. There is also a lack of clear and consistent evaluation criteria. Our survey aims to offer a comprehensive overview of current LLM planners to fill this gap. It builds on foundational work by Kartam and Wilkins (1990) and examines six key performance criteria: completeness, executability, optimality, representation, generalization, and efficiency. For each, we provide a thorough analysis of representative works and highlight their strengths and weaknesses. Our paper also identifies crucial future directions, making it a valuable resource for both practitioners and newcomers interested in leveraging LLM planning to support agentic workflows.\footnote{A curated list of papers and resources related to this survey are available at \url{https://github.com/wll199566/Awesome-LLM-Planning-Capability}.}

\end{abstract}

\tikzstyle{my-box}= [
    rectangle,
    draw=hidden-draw,
    rounded corners,
    text opacity=1,
    minimum height=1.5em,
    minimum width=5em,
    inner sep=2pt,
    align=center,
    fill opacity=.5,
]
\tikzstyle{leaf}=[my-box, minimum height=1.5em,
    fill=blue!15, text=black, align=left,font=\large,
    inner xsep=2pt,
    inner ysep
=4pt,
]
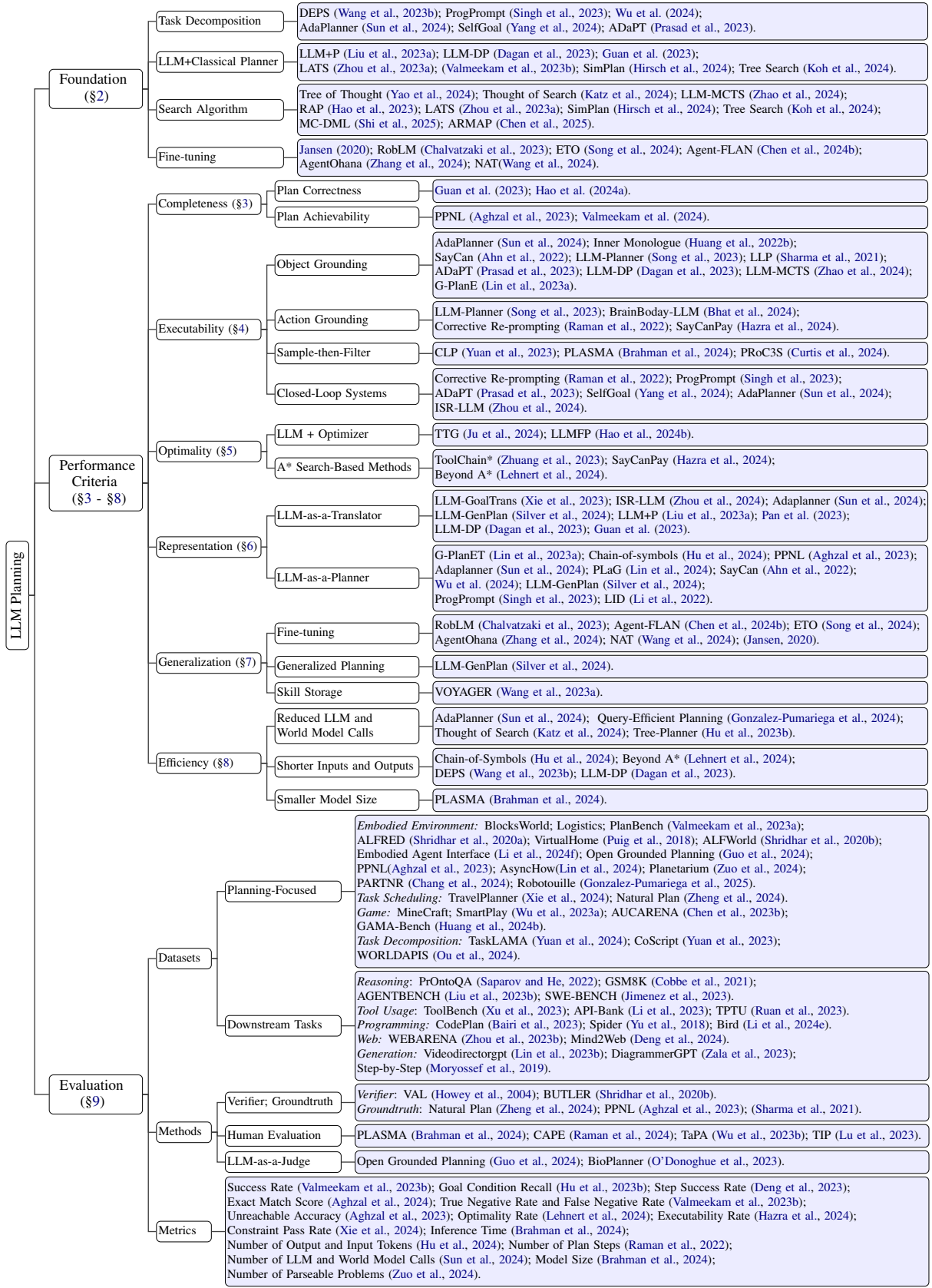
\begin{figure*}[htbp!]
    \centering
    \resizebox{\textwidth}{!}{
        \begin{forest}
            forked edges,
            for tree={
                grow=east,
                reversed=true,
                anchor=base west,
                parent anchor=east,
                child anchor=west,
                base=left,
                font=\Large,
                rectangle,
                draw=hidden-draw,
                rounded corners,
                align=left,
                minimum width=4em,
                edge+={darkgray, line width=1pt},
                s sep=3pt,
                inner xsep=2pt,
                inner ysep=3pt,
                ver/.style={rotate=90, child anchor=north, parent anchor=south, anchor=center},
            },
            where level=1{text width=7.9em,font=\Large,, align=left}{},
            where level=2{text width=8.9em,font=\large,}{},
            where level=3{text width=6.4em,font=\large,}{},
            where level=4{text width=6.4em,font=\large,}{},
        [ \;\;LLM Planning\;\; , ver
                [\;\;Foundation\\\;\;\;\;\;\;\;\;(\S \ref{sec:foundations})
                    [Task Decomposition,text width=10.9em
                        [
                            DEPS \cite{wang2023describe}; ProgPrompt \cite{singh2023progprompt}; \citet{wu2024can}; \\AdaPlanner \cite{sun2024adaplanner}; SelfGoal \cite{yang2024selfgoal}; ADaPT \cite{prasad2023adapt}.
                        ,leaf ,text width=55.5em]
                    ]
                    [LLM+Classical Planner ,text width=10.9em
                        [
                            LLM+P \cite{liu2023llm+}; LLM-DP \cite{dagan2023dynamic};  \citet{guan2023leveraging}; \\ LATS \cite{zhou2023language}; \cite{valmeekam2023planning}; SimPlan \cite{hirsch2024s}; Tree Search \cite{koh2024tree}.  
                        ,leaf ,text width=55.5em]
                    ]
                    [Search Algorithm,text width=10.9em
                        [
                            Tree of Thought \cite{yao2024tree}; Thought of Search \cite{katz2024thought}; LLM-MCTS \cite{zhao2024large}; \\ RAP \cite{hao2023reasoning}; LATS \cite{zhou2023language}; SimPlan \cite{hirsch2024s}; Tree Search \cite{koh2024tree}; \\MC-DML \cite{shimonte2025}; ARMAP \cite{chenautonomous2025}.
                        ,leaf ,text width=55.5em]
                    ]
                    [Fine-tuning,text width=10.9em
                        [
                            \citet{jansen2020visually}; RobLM \cite{chalvatzaki2023learning};  ETO \cite{song2024trial}; Agent-FLAN \cite{chen2024agent}; \\ AgentOhana \cite{zhang2024agentohana}; NAT\cite{wang2024learning}.
                        ,leaf ,text width=55.5em]
                    ]
                ]
                [\;\;Performance \\ \;\;\;\;\;Criteria \\\;\;\;\;\;(\S \ref{sec:completeness} - \S\ref{sec:efficiency})
                    [Completeness (\S \ref{sec:completeness})                      
                        [Plan Correctness ,text width=12.4em
                            [\citet{guan2023leveraging}; \citet{hao2024large}.
                            , leaf, text width= 43.5em]
                        ]
                        [Plan Achievability,text width=12.4em
                            [PPNL \cite{aghzal2023can}; \citet{valmeekam2024llms}.
                            ,leaf, text width = 43.5em]
                        ]
                    ]
                    [Executability (\S \ref{sec:executability})
                        [Object Grounding,text width=12.4em
                            [AdaPlanner \cite{sun2024adaplanner}; Inner Monologue \cite{huang2022inner}; \\ SayCan \cite{ahn2022can}; LLM-Planner \cite{song2023llm}; LLP \cite{sharma2021skill}; \\ ADaPT \cite{prasad2023adapt}; LLM-DP \cite{dagan2023dynamic}; LLM-MCTS \cite{zhao2024large}; \\G-PlanE \cite{lin2023grounded}. 
                            ,leaf, text width=43.5em]
                        ]
                        [Action Grounding,text width=12.4em
                            [LLM-Planner \cite{song2023llm};  BrainBoday-LLM \cite{bhat2024grounding}; \\Corrective Re-prompting \cite{raman2022planning}; SayCanPay \cite{hazra2024saycanpay}.
                            ,leaf, text width= 43.5em]
                        ]
                        [Sample-then-Filter,text width=12.4em
                            [CLP \cite{yuan2023distilling}; PLASMA \cite{brahmanplasma}; PRoC3S \cite{curtis2024trust}.
                            ,leaf, text width= 43.5em]
                        ]
                        [Closed-Loop Systems,text width=12.4em
                            [Corrective Re-prompting \cite{raman2022planning}; ProgPrompt \cite{singh2023progprompt}; \\ ADaPT \cite{prasad2023adapt}; SelfGoal \cite{yang2024selfgoal};
                            AdaPlanner \cite{sun2024adaplanner}; \\ISR-LLM \cite{zhou2024isr}.
                            ,leaf, text width= 43.5em]
                        ]   
                    ]
                    [Optimality (\S \ref{sec:optimality})
                        [LLM + Optimizer,text width=12.4em
                            [TTG \cite{ju2024globe}; LLMFP \cite{hao2024planning}.  
                            ,leaf, text width= 43.5em]
                        ]
                        [A* Search-Based Methods,text width=12.4em
                            [ToolChain* \cite{zhuang2023toolchain}; SayCanPay \cite{hazra2024saycanpay}; \\Beyond A* \cite{lehnert2024beyond}.
                            ,leaf, text width= 43.5em]
                        ]
                    ] 
                    [Representation (\S \ref{sec:representation})
                        [LLM-as-a-Translator,text width=12.4em
                            [LLM-GoalTrans \cite{xie2023translating}; ISR-LLM \cite{zhou2024isr}; Adaplanner \cite{sun2024adaplanner}; \\LLM-GenPlan \cite{silver2024generalized}; LLM+P \cite{liu2023llm+}; \citet{pan2023data}; \\ LLM-DP \cite{dagan2023dynamic};  \citet{guan2023leveraging}.
                            ,leaf, text width= 43.5em]
                        ]
                        [LLM-as-a-Planner,text width=12.4em
                            [G-PlanET \cite{lin2023grounded}; Chain-of-symbols \cite{hu2024chain}; PPNL \cite{aghzal2023can}; \\Adaplanner \cite{sun2024adaplanner}; PLaG \cite{lin2024graph}; SayCan \cite{ahn2022can}; \\ \citet{wu2024can}; 
                            LLM-GenPlan \cite{silver2024generalized}; \\ ProgPrompt \cite{singh2023progprompt}; LID \cite{li2022pre}. 
                            ,leaf, text width= 43.5em]
                        ]
                    ]
                    [Generalization (\S \ref{sec:generalization})
                        [Fine-tuning,text width=12.4em
                            [
                                 RobLM \cite{chalvatzaki2023learning}; Agent-FLAN \cite{chen2024agent}; ETO \cite{song2024trial};\\ AgentOhana \cite{zhang2024agentohana}; 
                                NAT \cite{wang2024learning}; \cite{jansen2020visually}.
                            ,leaf ,text width= 43.5em]
                        ]
                        [Generalized Planning,text width=12.4em
                            [
                            LLM-GenPlan \cite{silver2024generalized}.
                            ,leaf ,text width= 43.5em]
                        ]
                        [Skill Storage,text width=12.4em
                            [VOYAGER \cite{wang2023voyager}.
                            ,leaf ,text width= 43.5em]
                        ]
                    ]
                    [Efficiency (\S \ref{sec:efficiency})
                        [Reduced LLM and \\ World Model Calls,text width=12.4em
                            [AdaPlanner \cite{sun2024adaplanner};  \space Query-Efficient Planning \cite{gonzalez2024query}; \\ Thought of Search \cite{katz2024thought}; Tree-Planner \cite{hu2023tree}.
                            ,leaf ,text width= 43.5em]
                        ]
                        [Shorter Inputs and Outputs,text width=12.4em
                            [Chain-of-Symbols \cite{hu2024chain}; Beyond A* \cite{lehnert2024beyond}; \\ DEPS \cite{wang2023describe}; LLM-DP \cite{dagan2023dynamic}. 
                            ,leaf ,text width= 43.5em]
                        ]
                        [Smaller Model Size,text width=12.4em
                            [PLASMA \cite{brahmanplasma}.
                            ,leaf ,text width= 43.5em]
                        ]
                    ]
                ]
                [\;\;Evaluation \\\;\;\;\;\;\;\;(\S \ref{sec:evaluation})
                    [Datasets, text width=4.5em
                        [Planning-Focused,text width=9.9em
                            [\emph{Embodied Environment:} BlocksWorld; Logistics; PlanBench \cite{valmeekam2023planbench}; \\ ALFRED \cite{shridhar2020alfred}; VirtualHome \cite{puig2018virtualhome}; ALFWorld \cite{shridhar2020alfworld}; \\Embodied Agent Interface \cite{li2024embodied}; Open Grounded Planning \cite{guo2024opengrounded}; \\PPNL\cite{aghzal2023can}; AsyncHow\cite{lin2024graph}; Planetarium \cite{zuo2024planetarium}; \\ PARTNR \cite{chang2024partnr}; Robotouille \cite{gonzalez2025robotouille}.\\ 
                            \emph{Task Scheduling:} TravelPlanner \cite{xie2024travelplanner};  Natural Plan \cite{zheng2024natural}. \\ 
                            \emph{Game:}  MineCraft; SmartPlay \cite{wu2023smartplay}; AUCARENA \cite{chen2023put}; \\ GAMA-Bench \cite{huang2024far}. \\  
                            \emph{Task Decomposition:} TaskLAMA \cite{yuan2024tasklama}; CoScript \cite{yuan2023distilling}; \\WORLDAPIS \cite{ou2024worldapis}.
                            ,leaf ,text width=50.5em]  
                        ]
                        [Downstream Tasks,text width=9.9em
                            [\emph{Reasoning}: PrOntoQA \cite{saparov2022language}; GSM8K \cite{cobbe2021training}; \\ AGENTBENCH \cite{liu2023agentbench}; SWE-BENCH \cite{jimenez2023swe}. 
                            \\ \emph{Tool Usage}: ToolBench \cite{xu2023tool}; API-Bank \cite{li2023api}; TPTU \cite{ruan2023tptulargelanguagemodelbased}. 
                            \\ \emph{Programming:} CodePlan \cite{bairi2023codeplanrepositorylevelcodingusing}; Spider \cite{yu2018spider}; Bird \cite{li2024can}.\\ 
                            \emph{Web:} WEBARENA \cite{zhou2023webarena}; Mind2Web \cite{deng2024mind2web}. \\ 
                            \emph{Generation:} Videodirectorgpt \cite{lin2023videodirectorgpt}; DiagrammerGPT \cite{zala2023diagrammergpt}; \\Step-by-Step \cite{moryossef2019step}.
                            ,leaf ,text width=50.5em]
                        ]  
                    ]
                    [Methods, text width=4.5em
                        [Verifier; Groundtruth,text width=9.9em
                            [\emph{Verifier}: VAL \cite{howey2004val}; BUTLER \cite{shridhar2020alfworld}. \\
                             \emph{Groundtruth}: Natural Plan \cite{zheng2024natural}; PPNL \cite{aghzal2023can}; \cite{sharma2021skill}. 
                            ,leaf ,text width=50.5em]
                        ]
                        [Human Evaluation,text width=9.9em
                            [PLASMA \cite{brahmanplasma}; CAPE \cite{raman2024cape}; TaPA \cite{wu2023embodied}; TIP \cite{lu2023multimodal}. 
                            ,leaf ,text width=50.5em]
                        ]
                        [LLM-as-a-Judge,text width=9.9em
                            [Open Grounded Planning \cite{guo2024opengrounded}; BioPlanner \cite{o2023bioplanner}.
                            ,leaf ,text width=50.5em]
                        ]
                    ]
                    [Metrics, text width=4.5em
                        [Success Rate \cite{valmeekam2023planning}; Goal Condition Recall \cite{hu2023tree}; Step Success Rate \cite{deng2023mind2web}; \\ Exact Match Score \cite{aghzal2024look}; True Negative Rate and False Negative Rate \cite{valmeekam2023planning}; \\ Unreachable Accuracy \cite{aghzal2023can}; Optimality Rate \cite{lehnert2024beyond}; Executability Rate \cite{hazra2024saycanpay};  \\ Constraint Pass Rate \cite{xie2024travelplanner}; Inference Time \cite{brahmanplasma}; \\ Number of Output and Input Tokens \citep{hu2024chain}; Number of Plan Steps \cite{raman2022planning}; \\Number of LLM and World Model Calls \cite{sun2024adaplanner};  Model Size \cite{brahmanplasma}; \\ Number of Parseable Problems \cite{zuo2024planetarium}.
                        ,leaf ,text width=62em]
                    ]
                ]               
        ]
        \end{forest}
    }
    \caption{Taxonomy of LLM Planning}
    \label{fig:taxonomy}
\end{figure*}

\section{Introduction}

Planning, which involves generating a sequence of actions to reach a desired goal state \cite{newell1958elements, kartam1990towards}, is fundamental to human intelligence. For example, when planning a trip to San Francisco, one would search for flights, book tickets based on budget and schedule, arrange local transportation to the airport, and consider alternatives in case of cancellations. These planning tasks require complex reasoning, world knowledge, decision-making, and the ability to adapt, making them a significant challenge for humans. To date, there has been a growing focus on developing LLM planners to automate these complex tasks.

A comprehensive survey of LLM planners would significantly propel research in this field. Prior studies have explored planning methods and evaluation benchmarks \cite{huang2024understanding, li2024lasp}. \citet{huang2024understanding} categorized planning methods into decomposition, plan selection, external modules, reflection, and memory, while \citet{li2024lasp} reviewed evaluation benchmarks across various domains. However, many of these benchmarks and systems are tailored to specific problems, making it hard to compare LLM planners across domains or determine the best planner for new tasks. Further, there is a lack of clear and consistent evaluation criteria. We believe this gap may hinder the development of advanced LLM planners.

Our survey builds on the foundational work of \citet{kartam1990towards} to address key evaluation criteria for LLM planners. 
The original paper highlighted challenges in evaluating early AI planning systems, which relied on heuristics and were confined to research labs. The initial criteria were categorized into performance, representation, and communication issues.
With more advanced LLM planning, we reexamine this critical framework and focus on six key evaluation criteria: \emph{completeness}, \emph{executability}, \emph{optimality}, \emph{representation}, \emph{generalization}, and \emph{efficiency}. For each criterion, we provide a thorough analysis of representative works, highlighting their strengths and weaknesses.

We contribute to the literature by addressing key research questions in LLM planning: What foundational capabilities distinguish them from earlier AI planners? How can we comprehensively measure their performance? We examine the datasets, evaluation methods, and metrics available to the community. We also highlight crucial areas where research is still lacking, including representation, hallucination, alignment, multi-agent planning, connections to agentic workflows, aiming to fill these gaps and advance the field. Figure \ref{fig:taxonomy} presents a taxonomy of six key performance criteria and representative techniques. For those new to LLM planning, we recommend a thorough read, while experts can focus on specific sections. Each section offers clear definitions, relevant works, and includes links to tables in the Appendix. We will dive into the details in the following sections.

\section{LLM Planning Foundations (Tables \ref{tab:foundation_1}-\ref{tab:foundation_3})} \label{sec:foundations} 

We begin by exploring LLM planning foundations, covering widely-used paradigms to provide background for readers unfamiliar with the field. It is broken down into four parts.

\vspace{-0.1in}
\paragraph{Task Decomposition} Task decomposition breaks down abstract goals into specific, manageable sub-goals. It helps mitigate errors by enabling verification at each step and makes LLM reasoning more tractable by narrowing the knowledge space.

Task decomposition can be performed \emph{sequentially, in parallel, or asynchronously}. Specifically, sequential decomposition \cite{wang2023describe, singh2023progprompt, sun2024adaplanner, wu2024can} 
requires that the precondition of the subsequent subgoal is the effect of the preceding subgoal. 
In contrast, parallel decomposition \cite{yang2024selfgoal} involves subgoals that share the \emph{same} precondition and effect, where achieving the final goal requires completing only \emph{one} of these subgoals. Asynchronous decomposition \cite{lin2024graph} involves parallelizing subgoals as well. However, these subgoals in distinct branches have \emph{unique} preconditions and effects. Asynchronous decomposition requires the completion of \emph{all} subgoals to achieve the overall goal. 

Moreover, task decomposition can be performed \emph{recursively}, applying any of the above three approaches at each step. For example, \citet{prasad2023adapt} recursively break down the goal until each subgoal can be easily executed in the environment.  

\vspace{-0.1in}
\paragraph{LLM + Classical Planner} Studies \cite{valmeekam2023planning, kambhampati2024can, kambhampati2024llms} show that LLMs struggle with independent planning. Classical planners, such as Fast Downward \cite{helmert2006fast}, ensure correct plans but depend on experts to translate user queries into formal representations, limiting scalability. A hybrid approach integrating LLMs with classical planners combines the world knowledge of LLMs with the precision and reliability of classical methods, addressing their individual limitations.

When integrated with classical planners, LLMs \emph{translate natural language problems into formal representations} or \emph{generate initial plans}. For example, LLM+P \cite{liu2023llm+}, LLM-DP \cite{dagan2023dynamic}, and \citet{guan2023leveraging} use LLMs to convert planning problems into PDDL  \cite{McDermott1998PDDLthePD}, solved by Fast Downward or BFS(f) \cite{lipovetzky2014width}. \citet{valmeekam2023planning} employs LLMs to generate an initial plan, guiding the LPG planner \cite{gerevini2002lpg}, which iteratively refines it until a correct solution is found.  

\vspace{-0.1in}
\paragraph{Search Algorithm} Search algorithms, including \emph{Breadth-First Search, Depth-First Search} \cite{yao2024tree, katz2024thought}, \emph{Monte Carlo Tree Search} \cite{hao2023reasoning, zhou2023language, zhao2024large, shimonte2025}, and \emph{Greedy Best-First Search} \cite{koh2024tree, hirsch2024s}, have been applied to improve LLM-based planning. These algorithms treat planning as a search problem, using search policy to guide the exploration of various possibilities. Search algorithms excel in planning problems by offering systematic exploration, optimality guarantees, and formal verification without requiring extensive domain knowledge, though they may be computationally intensive compared to more specialized methods like task decomposition.

All search algorithms consist of four core components: (1) \emph{Search Policy} determines node exploration order, which are defined by the underlying search algorithm and are independent of LLMs. (2) \emph{Expansion} generates possible actions from a state, often using LLMs to propose actions based on user instructions and current environment. (3) \emph{World Models} define state transitions based on action preconditions and effects, using LLMs \cite{hao2023reasoning}, classical planners \cite{hirsch2024s}, or external environment simulators \cite{zhou2023language, zhao2024large, koh2024tree}. (4) \emph{Evaluation} assesses state progress toward the goal via scores computed by predefined functions \cite{katz2024thought}, LLM/LVM ratings \cite{yao2024tree, hao2023reasoning, zhou2023language}, log-likelihood scores \cite{hirsch2024s}, voting \cite{yao2024tree}, self-consistency scores \cite{zhou2023language} or reward models \cite{chenautonomous2025}. 

\vspace{-0.1in}
\paragraph{Fine-tuning} 

Current LLMs are not specifically trained for agentic tasks like planning, and prompt-based methods, which do not update model parameters, cannot fundamentally improve performance in these areas \cite{chen2023fireact, wang2024learning}. Fine-tuning, either focused on \emph{planning-specific tasks} or \emph{broader agentic capabilities}, enhances planning correctness by directly updating LLM parameters.

Planning-specific fine-tuning involves training a pretrained model on planning-focused tasks (e.g., Blocksworld or ALFWorld \cite{shridhar2020alfworld}), to improve planning performance. For example, \citet{jansen2020visually} and \citet{chalvatzaki2023learning} fine-tuned GPT-2 \cite{radford2019language} on ALFWorld, demonstrating its effectiveness in robotics planning. 

Generalized agentic fine-tuning optimizes models using datasets that include both general tasks (e.g., question answering) and diverse agentic tasks (e.g., reasoning, planning, and tool use). This approach is motivated by two key insights: (1) focusing too narrowly on planning may degrade general capabilities \cite{chen2024agent}, and (2) agentic tasks share overlapping capabilities. For instance, reasoning and tool-use tasks often involve planning components. Thus, fine-tuning on a broader set of agentic tasks can simultaneously enhance planning performance and other interrelated capabilities, like reasoning, which are integral to planning. Moreover, standardizing trajectory formats from different tasks \cite{zhang2024agentohana, chen2024agent}, as well as incorporating unsuccessful reasoning or planning trajectories \cite{wang2024learning, chen2024agent, song2024trial} can further enhance learning and performance.

\section{Criterion I: Completeness (Table \ref{tab:completeness})}\label{sec:completeness}

The completeness of planning has two key aspects: (1) if a valid plan exists, the model should generate it correctly, and (2) if no feasible plan is possible, the model should recognize this and refrain from generating an incorrect or arbitrary plan.

A plan is correct if it achieves the goal within a fixed budget while avoiding excessive complexity and infinite loops. To ensure correctness, the LLM must work with classical sound and complete solvers \cite{guan2023leveraging, hao2024large}. Also, the LLM has to accurately translate the domain and problem into the specific format (e.g., PDDL), required by these solvers \cite{guan2023leveraging}.

In terms of identifying unsolvable planning problems, those with inherently unachievable goals, even top LLMs (e.g., GPT-4 \cite{achiam2023gpt}) and Large Reasoning Models (e.g., OpenAI O1 \cite{jaech2024openai}) struggle due to hallucination issues \cite{aghzal2023can, valmeekam2024llms, katz2024thought}.

\section{Criterion II: Executability (Tables \ref{tab:executability_1}-\ref{tab:executability_2})} \label{sec:executability}

Executability checks if a plan can be carried out in a given environment while meeting all constraints. A executable plan must use only allowed actions and recognizable objects. Beyond basic precondition and postcondition rules, planners must consider extra constraints, such as avoiding sugar when baking a cake for diabetics \cite{yuan2023distilling}. Importantly, executability and correctness are orthogonal: \emph{an executable plan isn’t necessarily correct}, since it might be grounded and follow all constraints but still fail to reach the goal; likewise, \emph{a correct plan isn’t always executable} since it may only include high-level steps that can’t be executed in a specific environment. Real-world applications typically require plans that are both correct and executable, especially when the executors are not humans (e.g., robots and computers).

To ensure plans are executable, researchers have proposed several approaches, including Object Grounding, Action Grounding, Closed-Loop Systems, and Sample-then-Filter. \smallskip

\noindent\textbf{Object Grounding} Object grounding ensures the LLM planner uses objects available in the current environment when generating plans. For example, if a stove is available but a microwave is not, the planner should choose the stove to heat a pancake and disregard the unavailable microwave in its plan.
The simplest way to do this is by feeding observed or available objects into the planner via prompts \citep{huang2022inner, song2023llm, lin2023grounded, singh2023progprompt} or neural embeddings \cite{sharma2021skill, ahn2022can}. In partially observed environments, where some object information are uncertain (e.g., needing to clean a cup that could be in a cabinet, drawer, or fridge), the planner can generate multiple possible plans, one for each scenario, and select the first feasible one \citep{prasad2023adapt, dagan2023dynamic, zhao2024large}. \citet{sun2024adaplanner} takes a different approach, first generating a plan with placeholders for objects, then filling in the blanks with observed objects during execution.\smallskip

\noindent\textbf{Action Grounding} Action grounding ensures all actions in a plan can actually be executed in the current environment. E.g., the planner should avoid including actions such as `draw a star' if the robot arm cannot perform such a complex operation. Instead, the task should be decomposed into simpler, supported actions, such as multiple steps of 'draw a line'. Like object grounding, the simplest way is to explicitly list all admissible actions in LLMs’ inputs \cite{singh2023progprompt}. If a step goes beyond the executor’s capabilities (e.g., combining multiple allowed actions into one), the LLM planner should be reprompted to break it down until every step is executable \citep{prasad2023adapt}. 

Hierarchical Planning is another common method for grounding actions in LLM planning \citep{huang2022language, raman2022planning, song2023llm, hazra2024saycanpay, bhat2024grounding}. It starts with high-level steps and then translates each one into a sequence of executable actions. This can be done in two ways: either generating all high-level steps first and then refining them into actions or translating each step as it's generated. If an action isn't exactly admissible, the closest valid action is retrieved instead \citep{huang2022language, raman2022planning}. \smallskip

\noindent\textbf{Sample-then-Filter} Since LLMs alone can't guarantee plans meet all constraints, this approach first generates multiple plans and then verifies them, selecting only those that pass all checks. \citet{yuan2023distilling} ranks InstructGPT-generated plans using cosine similarity with task embeddings and selects the most similar one. \citet{brahmanplasma} applies a verifier-guided beam search, keeping the top-K plans based on correctness and constraint adherence at each step. \citet{curtis2024trust} generates Pythonic plans with parameter ranges, tests them with a simulator or classifier, and prompts the LLM to revise if constraints are still violated. \smallskip

\noindent\textbf{Closed-Loop Systems}\;\; A closed-loop system in LLM planning means the model adapts its plan based on feedback from executors \cite{prasad2023adapt, yang2024selfgoal}, simulators \cite{bhat2024grounding}, validators \cite{zhou2024isr, silver2024generalized}, other LLMs \cite{wang2023describe, zhou2023language}, or even humans \cite{huang2022inner}, when the initial plan are inexecutable. It reprompts the LLM planner to replan 
until the plan is fully executable. Unlike open-loop systems \cite{huang2022language}, which lack feedback, closed-loop planning helps reduce hallucinations and enables LLMs to handle complex, long-horizon, and dynamic environments \cite{wang2023describe}.

Closed-loop systems fall into two types: \emph{implicit} and \emph{explicit} \cite{sun2024adaplanner}. Implicit systems only fix the failed action \cite{raman2022planning, singh2023progprompt, zhou2024isr, prasad2023adapt, yang2024selfgoal}, while explicit systems regenerate the entire plan \cite{sun2024adaplanner}. Though explicit systems require more computation, they prevent errors from compounding across steps.

\section{Criterion III: Optimality (Table \ref{tab:optimality})} \label{sec:optimality}

Optimality means achieving the goal state through the \emph{best} possible plan. It poses a greater challenge than standard planning, which only requires reaching the goal state. Researchers have proposed two paradigms for achieving the optimal plans: LLM + Optimizer and $\textbf{A}^*$ search-based methods. \smallskip

\noindent\textbf{LLM + Optimizer}\;\; It combines the LLM, which turns user requests into symbolic optimization problems, with an optimizer that solves them and finds the best solution \cite{ju2024globe, hao2024planning}. For example, TTG \citep{ju2024globe} uses the LLM to convert travel planning requests of minimum total costs into Mixed Integer Linear Programming problems, then runs an optimizer such as SCIP \citep{bestuzheva2021scip} to provide the optimal plan. LLM + Optimizers ensure optimal solutions by leveraging LLMs to formulate constrained optimization problems and classical optimizers to solve them. In contrast to classical planners, which typically rely on search algorithms, heuristics, and logical deduction and may not guarantee optimality \cite{russell2016artificial}, optimizers, often based on gradient-based methods (e.g., Newton's methods), can \emph{guarantee optimal solutions} \cite{Boyd_Vandenberghe_2004}.\smallskip

\noindent\textbf{A* Search-Based Methods}\; A* search finds the lowest-cost optimal solution, particularly when using admissible heuristics that do not overestimate the actual cost to the goal. This makes it a natural choice for LLM-based planners to achieve optimality. ToolChain* \cite{zhuang2023toolchain} combines A* tree search with an LLM, which suggests next steps and estimates heuristic scores, to create plans with the fewest tool API calls. SayCanPay \citep{hazra2024saycanpay} uses A* search with LLMs to generate the shortest possible plans. Beyond A* \citep{lehnert2024beyond} trains a Transformer model, Searchformer, to mimic A* search paths for complex tasks like Maze navigation and Sokoban puzzles, optimizing for the fewest steps. Besides A* search, other search algorithms (e.g., DFS and MCTS) could also be used to find optimal solutions, although without guarantees.

\section{Criterion IV: Representation (Tab. \ref{tab:representation_1}-\ref{tab:representation_2})} \label{sec:representation}

In LLM planning, representation refers to how inputs and outputs are formatted. Inputs include domains (predicates and actions), problems (initial and goal states), and environmental observations, while outputs are the generated plans. Effective representation enhances problem comprehension and execution efficiency, especially given LLMs' sensitivity to prompts. We discuss this in two contexts: LLM-as-a-Translator and LLM-as-a-Planner. 

\vspace{-0.05in}
\paragraph{LLM-as-a-Translator} LLM-as-a-Translator converts between natural language (NL) and formal planning languages (e.g. PDDL), making classic planners more accessible to non-experts. By converting natural language tasks into formal representations and translating the resulting plans back into NL, LLMs reduce ambiguity, minimize hallucinations, and enable external validation, improving both usability and reliability in planning systems \cite{xie2023translating, zhou2024isr, sun2024adaplanner, silver2024generalized}. 

Recent work has used LLMs to translate natural language descriptions into PDDL \cite{liu2023llm+, guan2023leveraging, xie2023translating, dagan2023dynamic, zhou2024isr, tantakoun2025llmsplanningmodelerssurvey}, LTL \cite{pan2023data}, and STL \cite{chen2024autotamp}. To ensure reliability, translations should be tested on development or external datasets like Planetarium \cite{zuo2024planetarium}. If there are syntax or semantic errors, validators (e.g. VAL \cite{howey2004val}) or human experts can provide feedback for the LLM to fix them. 

\vspace{-0.05in}
\paragraph{LLM-as-a-Planner} When LLMs act as standalone planners without classical planners or optimizers, various methods help encode environmental information, domains, and plans beyond just natural language. Environment and domain details have been represented using \emph{tables} \cite{lin2023grounded}, \emph{condensed symbols} \cite{hu2024chain}, \emph{Pythonic code} \cite{aghzal2023can, singh2023progprompt, sun2024adaplanner}, \emph{neural embeddings} \cite{li2022pre, ahn2022can}, and \emph{graphs} \cite{lin2024graph, wu2024can}. For generated plans, Pythonic code is a common alternative to natural language \cite{singh2023progprompt, silver2024generalized}.

\section{Criterion V: Generalization (Table \ref{tab:generalization})} \label{sec:generalization}

Generalization refers to LLM planners' ability to apply learned strategies to new, more complex out-of-domain scenarios beyond its training environment, which can be enhanced through three key approaches: \emph{fine-tuning} (described previously in Section \ref{sec:foundations}), \emph{generalized planning}, and \emph{skill storage}. Given the diverse user queries in the real-world deployments, ensuring LLM planners' generalizability is important alongside other performance.

\vspace{-0.05in}
\paragraph {Generalized Planning} Generalized planning extracts common patterns from a limited set of training solutions (i.e., plans) to solve unseen tasks within the same domain, which may be larger and more complex than the training tasks \cite{srivastava2011new}. 
For example, in the Delivery dataset \cite{yang2022pg3}, models trained on small-scale deliveries (9–17 locations) can generalize to larger ones (70–100 locations) using the same core strategy. 
\citet{silver2024generalized} approached this by prompting LLMs to summarize the domain and generate a minimal, generalizable Python-based plan.  

\vspace{-0.05in}
\paragraph {Skill Storage} Skill storage focuses on learning and reusing previously acquired skills to tackle new problems. E.g., \citet{wang2023voyager} introduced a skill library that stores successfully executed skills (e.g., Combat Zombie). These skills are abstracted and generalized for reuse in similar situations (e.g., fighting spiders involves similar actions to fighting zombies). When encountering an unseen task, the LLM planning system retrieves relevant learned skills based on the task and current states, then applies them to generate an effective solution.

\section{Criterion VI: Efficiency (Table \ref{tab:efficiency})} 
\label{sec:efficiency}

Efficiency in LLM planning means reducing computational and monetary costs by decreasing LLM calls, world model interactions, input and output lengths, and model sizes. This is crucial especially developing planners based on commercial LLMs.

\vspace{-0.08in}
 \paragraph{Reduced LLM and World Model Calls}\;\; To reduce the number of LLM and world model calls, several tricks are used: (1) generating the entire plan in one shot instead of step-by-step to reduce redundant prompts \cite{hu2023tree, sun2024adaplanner, gonzalez2024query}; (2) checking feasibility by world models only at the end of each subgoal, not after every action \citep{sun2024adaplanner, gonzalez2024query}; (3) merging plans with the same prefix actions or subgoals to avoid duplicate world model checks when sampling multiple plans \citep{hu2023tree}; and (4) in tree search-based methods, querying the LLM once to generate a successor function and a goal-check function. The successor function lists all possible actions and resulting states based on the current state, while the goal-check function determines if a state meets the goal. This approach avoids repeated LLM and world model calls at each node \citep{katz2024thought}.  

\vspace{-0.05in}
\paragraph{Shorter Inputs and Outputs} Reducing input and output length includes \emph{decreasing prompt and plan tokens} and \emph{minimizing actions in the final plan} to alleviate the load on executors. For spatial reasoning and planning, \citep{hu2024chain} introduces Chain-of-Symbols (CoS), a compact symbolic representation that replaces natural language descriptions in CoT \cite{wei2022chain} trajectories. \citet{lehnert2024beyond} uses search dynamic bootstrapping to iteratively fine-tune a LLM, replacing training cases with solutions with less tokens and equal optimality. To minimize actions, \citet{dagan2023dynamic} and \citet{wang2023describe} use action selectors based on predefined rules or trained models to find the shortest successful plan.     

\vspace{-0.05in}
\paragraph{Smaller Model Sizes} Shrinking the model size can reduce the computational burden, accelerating training and inference while lowering costs. To train a smaller planning model, \citet{brahmanplasma} uses GPT-3 \cite{brown2020language} as the teacher and T5 \cite{raffel2020exploring} as the student, distilling the teacher's planning capabilities into the more compact student model.

\section{Evaluation} \label{sec:evaluation}

\vspace{-0.05in}
 \paragraph{Datasets}  LLM planning evaluation is conducted on two types of datasets: \emph{planning-focused datasets} and \emph{downstream-task datasets}. 

Planning-focused datasets primarily assess planning abilities. The most common scenarios include (1) \emph{Embodied environments}, (2) \emph{Task scheduling} , (3) \emph{Games}, and (4) \emph{Task decomposition} \cite{li2024lasp}. Figure \ref{fig:taxonomy} presents commonly used planning datasets; readers can refer to \citet{li2024lasp,li2025planetcollectionbenchmarksevaluating} for further details.

While most of the datasets mentioned above assess whether the generated plans are correct, some specifically target key performance criteria in LLM planning. For \emph{grounding}, Open Grounded Planning \cite{guo2024opengrounded} and Embodied Agent Interface \cite{li2024embodied} evaluate performance in embodied environments, while CoScript \cite{yuan2023distilling}, TravelPlanner \cite{xie2023translating}, and PPNL \cite{aghzal2023can} focus on planning problems with constraints. For \emph{representation}, Planetarium \cite{zuo2024planetarium} assesses LLMs' ability to translate natural language into PDDL. For \emph{optimality}, \citet{lin2024graph} and \citet{gonzalez2025robotouille} introduce tasks requiring optimal plans using asynchronous actions. PPNL \cite{aghzal2023can} can also evaluate a planner’s ability to \emph{identify unachievable goals (i.e., completeness)}.

Planning abilities can also be evaluated through downstream tasks, where planning is integral to task completion, and stronger planning skills enhance overall performance. Downstream tasks can be categorized as follows: (1) \emph{Agentic tasks}, including reasoning-oriented tasks, tool-use-oriented tasks, programming tasks, and web tasks \cite{yu2018spider, cobbe2021training, saparov2022language, zhou2023webarena, deng2023mind2web, liu2023agentbench, li2023api, xu2023tool, jimenez2023swe, ruan2023tptulargelanguagemodelbased, bairi2023codeplanrepositorylevelcodingusing, li2024can}, (2) \emph{Generation tasks}, including video \cite{lin2023videodirectorgpt}, image \cite{zala2023diagrammergpt} and text generation \cite{moryossef2019step}. Please refer to Figure \ref{fig:taxonomy} for example datasets.
 
\vspace{-0.08in}
\paragraph{Methods}
The most common approach to evaluating LLM planning is to test it in a simulated environment and validate the generated plans using either \emph{an internal verifier} provided by the environment or \emph{external verifier} (e.g., VAL \cite{howey2004val}) to ensure they achieve the intended goal. When ground-truth plans are available, LLM-generated plans can also be \emph{compared against these reference plans} \cite{zheng2024natural}.

The second evaluation method is \emph{human evaluation}, typically used in the following cases: (1) No available verifier: when certain simulated environments (e.g., VirtualHome) or real-world scenarios (e.g., using a mobile manipulator) lack automated verification; (2) Open-ended problems: tasks with ambiguous instructions or generative outputs (e.g., text or images) where multiple valid solutions may differ from the ground truth.

The final evaluation method, \emph{LLM-as-a-Judge}, uses another LLM to automatically assess the quality of generated plans in the cases mentioned above. This approach has been increasingly adopted in recent LLM planning research \cite{guo2024opengrounded, o2023bioplanner}. Compared to human evaluation, LLM judges are faster and more cost-effective, making them especially valuable for evaluating large datasets. However, this method has limitations, such as position bias, length bias, self-inconsistency, and sensitivity to prompts \cite{zheng2023judging, ye2024justice, wei2024systematic}. Addressing these issues is crucial to ensure reliable assessments. For more details on LLM-as-a-Judge, please see \citet{li2024generation, li2024llms, gu2024survey}.
 
\vspace{-0.05in}
\paragraph{Metrics}\;
Figure \ref{fig:taxonomy} summarizes commonly used evaluation metrics for planning-focused tasks, along with representative works. Performance criteria are measured using specific metrics: (1) \emph{Completeness}: success rate and goal condition recall measure whether the generated plan reaches final or stepwise goals, while classification metrics (e.g., true negative rate, false negative rate, and unreachable accuracy) assess the planner’s ability to identify unachievable tasks. When ground-truth plans are available, the exact match score is used. (2) \emph{Executability}: executability rate evaluates whether the plan can be executed in the environment, while constraint pass rate checks if constraints are met.  (3) \emph{Optimality}: measured by the optimality rate (i.e., the percentage of optimally solved tasks). (4) \emph{Efficiency}: common metrics include inference time, input and output token counts, number of plan steps, and model size.  (5) \emph{Representation}: the number of parseable problems indicates correct translations.  (6) \emph{Generalization}: all these metrics can also be applied to unseen scenarios to assess generalization. See Figure \ref{fig:taxonomy} for definitions of individual metrics and representative works.

\vspace{-0.04in}
\section{Discussion}
\label{sec:discussion}

In this section, we discuss the limitations in current LLM planning research studies, and suggest future directions for improvement and more comprehensive evaluations of LLM planning performance.

\vspace{-0.08in}
\paragraph{Datasets and Baselines} The current evaluation of LLM planning has its limitations, primarily because studies often rely on limited datasets and baselines. This makes it tough to fairly and comprehensively compare different methods. Most studies only use a few datasets from a single domain and difficulty level, and they do not evaluate all the six performance criteria. Inconsistent dataset choices make direct comparisons difficult. On top of that, many studies only compare against basic baselines such as CoT or ReAct, which does not help in comparing more advanced approaches. To fix this, a public, standardized leaderboard should be set up that covers all performance criteria, uses consistent evaluation metrics, includes a variety of baseline and advanced methods, and utilizes diverse datasets spanning multiple domains and difficulty levels. Another useful direction would be to create multilingual planning datasets and assess LLM performance across different languages.

\vspace{-0.08in}
\paragraph{Representation}\; LLMs are highly sensitive to prompts \cite{sclar2024quantifyinglanguagemodelssensitivity, razavi2025benchmarkingpromptsensitivitylarge}, but most research relies on natural language without comparing them to alternative formats, such as PDDL or Python, for describing domains and problems. Some studies \cite{singh2023progprompt, aghzal2024look} suggest that using Python to represent planning problems can improve performance, but automatically translating natural language problem descriptions into Python remains challenging, particularly for non-experts. If LLMs are to handle this translation effectively, additional datasets and evaluations are needed to assess their performance. Furthermore, little research has been conducted on how different prompt templates impact LLM planning performance, or on the best output formats for representing plans. Lastly, most fine-tuning methods rely on natural language data without exploring other formats, such as symbolic representations. Filling these gaps requires building benchmarks like Planetarium \cite{zuo2024planetarium} and carefully choosing representation formats in experiments.

\vspace{-0.08in}
\paragraph{Hallucination}  LLMs often experience hallucinations \cite{huang2023survey}, which present two major challenges in planning. First, they might struggle to assess if a plan is achievable given a specific problem description \cite{aghzal2023can, kambhampati2024can}. Second, they can generate inadmissible actions and non-existent objects, requiring translation or expert intervention to correct them \cite{huang2022language, raman2022planning}. This increases the cost of planning systems. Further research is needed to understand the root causes and improve LLMs' ability to accurately identify unachievable plans. Evaluating the impact of these hallucinations remains an important research direction.

\vspace{-0.08in}
\paragraph{Human Preference Alignment}\; There is a gap in understanding whether system generated plans align with human preferences. It is crucial for open-ended problems where humans execute the plans. Ensuring alignment with human preferences is vital for safety, feasibility, and usability, particularly in personalized planning tasks such as calendar and travel planning. Additionally, \citet{aghzal2024look} found that LLM planners often fail to achieve optimality in path planning, frequently producing unnecessarily long plans. This may stem from inherent length bias in LLMs, which tend to generate longer sequences. Alignment techniques such as RLHF \cite{ouyang2022training} and DPO \cite{rafailov2024direct} may help alleviate this issue, as humans generally prefer shorter plans for their efficiency, simplicity, and cognitive ease. Further investigation is needed to better align LLM planners with human preferences.

\vspace{-0.07in}
\paragraph{Cost Effectiveness} Current methods, particularly task decomposition and search-based approaches, often consume a large number of tokens due to lengthy prompts and repeated LLM queries. While heuristic search is considered more efficient than task decomposition, it still requires substantial repeated prompting. To improve cost-effectiveness, we may summarize problem descriptions and enhance heuristic evaluations, e.g., by improving LLM uncertainty estimation \cite{huang2024survey} and verification \cite{li2024systematicanalysisllmcontributions}. These improvements would help reduce prompt length and enable the early stopping of unpromising partial plans.

\vspace{-0.07in}
\paragraph{Multi-Agent Planning} Most existing research focuses on single-agent planning, where only one agent performs a task. Multi-agent planning \cite{konolige1980multiple, torreno2017cooperative} is more challenging, as it involves multiple agents (e.g., robots) working together or competing in parallel. Despite its complexity, multi-agent planning has received limited attention in AI planning research. It often requires coordinating multiple agents in collaborative or competitive environments where they operate simultaneously. The major challenge lies in developing effective communication protocols and cooperation strategies while generating viable plans for their collective actions.

\vspace{-0.07in}
\paragraph{Reasoning, Tool Use, and Memory}\quad There is often limited discussion on how other components of LLM agents, such as reasoning, tool use, and memory, affect planning performance. In particular, when LLMs are combined with classical planners or optimizers, it is crucial that the LLM accurately translates the planning problem into the appropriate domain representation to ensure correct plan generation. Currently, these approaches rely on human-selected planners and optimizers. Treating them as tools that LLMs can autonomously choose from could be an exciting prospect. This also raises the question of whether LLMs can effectively select the best tool for a given planning task. Future research should look into enhancing these agentic capabilities in LLM-based planning.

\section{Conclusion}
\label{sec:conclusion}

In our survey, we explore the landscape of modern LLM planners, proposing key performance criteria and discussing evaluation challenges. Our proposed criteria offer a structured approach to assess LLM planners across diverse domains. By systematically analyzing existing systems, datasets, and evaluation strategies, we aim to provide a foundation for future research in this space. We encourage researchers to build on our findings to create robust, highly adaptable, and efficient LLM planners.

\section{Limitations} \label{sec:limitation}

This work primarily focuses on commonly studied domains involving single-agent scenarios, such as robotics, household tasks, and computer-based tasks. We acknowledge that LLM planning is also applied in other areas, including the natural sciences \cite{o2023bioplanner, liu2024multimodal}, the Internet of Things \cite{cui2024llmind}, and multi-agent scenarios. However, these studies follow similar methodologies and evaluations, suggesting our survey's comprehensiveness. We focus on six commonly used performance criteria and exclude others, such as security and personalization, due to limited research in these areas. Instead, we discuss them in our future directions section.

\section*{Acknowledgements}

We thank the reviewers for their insightful feedback, which has helped improve our paper. H.W. and S.P. are supported by a UC Merced Spring 2023 Climate Action Seed Competition Grant, CAHSI-Google Institutional Research Program Award, and F3 R\&D GSR Award funded by the US Department of Commerce, Economic Development Administration Build Back Better Regional Challenge. Z.Z. and F.L. are partially supported by NSF CAREER Award \#2303655. 

\bibliography{custom}

\section{Appendix}
\label{sec:app_table}

\begin{table*}[!h]
\renewcommand{\arraystretch}{1.5} 
\small
\centering
\caption{Summary of Foundations in LLM Planning (Section \ref{sec:foundations})}
\vspace{-0.5em}
\scalebox{0.67}{
\begin{tabular}{|p{2.8cm}|p{2.2cm}|p{3.0cm}|p{5.5cm}|p{6.0cm}|}

\hline
\textbf{Method Name} & \textbf{Dataset} & \textbf{Evaluation Metric} & \textbf{Methods} & \textbf{Major Contribution} \\

\hline
DEPS \par \cite{wang2023describe} & Alfworld & Success Rate, Average Episode Length & 1. LLM generates PDDL goal from task description \par 2. Sampling world beliefs and using symbolic planning \par 3. Plan generator (BFS(f)) for generating plans & 1.LLM-DP achieves 96$\%$ success in Alfworld, outperforming the ReAct baseline (53$\%$) with fewer actions. \par 2.Enhances correctness in LLM planning by dynamically adjusting plans in response to environmental feedback and sub-goal feasibility.\\

\hline
ProgPrompt \par \cite{singh2023progprompt} & VirtualHome, Real-World Robot Tasks & Success Rate, Goal Conditions Recall, Executability & 1. Pythonic program generation for task planning \par  2. Use of natural language comments and assertions for feedback \par 3. Integration with real-time environment state feedback during task execution & 1. ProgPrompt significantly outperforms baseline methods by using programmatic LLM prompts to generate executable task plans, achieving up to 1.00 SR and 1.00 Exec in various VirtualHome tasks. It also adapts well to real-world robot tasks, with a Plan SR of 1 in most cases. \par 2. Improves correctness by incorporating environmental state feedback directly into the planning process. \\

\hline
Adaplanner \par \cite{sun2024adaplanner} & ALFWorld and MiniWoB++ & Success Rate & Closed-loop approach allowing LLM agent to refine self-generated plan adaptively & 1. Uses code-style LLM prompt structure and skill discovery mechanism. Achieves 91.79$\%$ success rate on ALFWorld tasks. \par 2. Achieves 91.11$\%$ success rate on MiniWoB++ tasks with feedback. \par 3. Improves planning correctness by adaptively refining plans based on environmental feedback, effectively managing complex sequential tasks. \\

\hline
GNN-Enhanced Task Planner \par \cite{wu2024can} & HuggingFace, Multimedia tasks, Daily Life API tasks, TMDB API tasks & Node F1-Score, Link F1-Score, Task Accuracy, Token Consumption & 1. Integration of GNNs for task graph navigation \par 2. Training-free (SGC) and training-required (GraphSAGE) approaches \par  3. GNN-based node and edge selection for task planning \par 4. Training fine-tuned models for better task retrieval & 1. GNN-enhanced planning outperforms LLM-based solutions by improving task accuracy (up to 9$\%$) with reduced token consumption. The proposed method scales well with larger task graphs and improves planning efficiency by a significant margin.\par 2.Addresses correctness by effectively mapping task dependencies in a graph structure, enhancing sub-task selection and sequence planning. \\

\hline
SelfGoal \par \cite{yang2024selfgoal} & Public Goods Game, Guess 2/3 of the Average, First-Price Auction, Bargaining & Success Rate, TrueSkill Score, Contribution Consistency & 1. Constructs GOALTREE to decompose high-level goals dynamically \par 2. Uses Search Module to select the most relevant subgoals \par 3. Decomposition updates based on environmental feedback \par 4. Adaptive subgoal tree refinement during task execution & 1. SelfGoal achieves a 94$\%$ success rate in dynamic multi-agent environments, outperforming baselines (e.g., ReAct, ADAPT) by dynamically adjusting subgoal guidance. It significantly improves task performance, especially in cooperative and competitive tasks.\par 2. SelfGoal enhances correctness and adaptability in planning by dynamically adjusting subgoals and actions based on ongoing task performance and feedback.\\

\hline
ADaPT \par \cite{prasad2023adapt} & ALFWorld, WebShop, TextCraft $\&$ hotel data & Success Rate, Task Complexity Handling, Sub-Task Decomposition Efficiency & 1. As-needed recursive decomposition of complex tasks \par  2. Planner and executor modules with dynamic failure adaptation \par 3. Multi-level decomposition for task complexity handling & ADaPT improves success rates by up to 33$\%$ over baselines, dynamically decomposing complex tasks as needed. It handles task complexity efficiently with up to 28.3$\%$ higher success rates on ALFWorld and 27$\%$ higher on WebShop.\\

\hline
LLM+P \par \cite{liu2023llm+} & Blocksworld, Grippers, Barman, Termes, and other robot planning scenarios & Success Rate, Optimal Plan Success, Plan Execution Time & 1. Converts natural language problem description into PDDL format \par 2. Uses classical planners (e.g., FAST-DOWNWARD) to generate optimal plans \par 3. Translates planner output back into natural language & 1. LLM+P significantly outperforms LLMs in solving long-horizon planning problems, achieving optimal plans in robot domains with high success rates (up to 100$\%$ on Blocksworld) and minimal execution time. 2. LLM+P improves correctness by integrating LLMs with classical planning techniques to generate and execute optimal plans.\\

\hline
LLM-DP \par \cite{dagan2023dynamic} & Alfworld & Success Rate, Average Episode Length & 1. LLM converts task descriptions into executable PDDL goals \par 2. Symbolic planner (BFS(f)) generates valid plans \par 3. Belief sampling to generate multiple world states for planning & LLM-DP outperforms the ReAct baseline, achieving 96$\%$ success in Alfworld, with significantly fewer actions (13.16 vs. 18.69) per task and higher efficiency due to belief sampling and symbolic planning integration.\\

\hline
PDDL World Model Generation \par \cite{guan2023leveraging} & Household, Tyreworld, Logistics & Error Count, Success Rate & 1. LLM-based PDDL generation for task planning \par 2. Error correction using LLMs as feedback interfaces \par 3. Use of external planners for generating feasible plans from PDDL models & 1. The paper shows that GPT-4 is capable of generating high-quality PDDL models with fewer errors than GPT-3.5-Turbo. It demonstrates that GPT-4-based world models lead to a 95$\%$ success rate in planning tasks using classical planners like Fast Downward. \par 2. Enhances plan correctness by using LLM-generated PDDL models in conjunction with domain-independent planners, ensuring plans are not only feasible but also optimal.\\

\hline
LLM Planning Evaluation \par \cite{valmeekam2023planning} & Blocksworld, Mystery Blocksworld & Success Rate, Levenshtein Edit Distance & 1. Evaluates LLMs (GPT-3, BLOOM) on common planning domains \par 2. Three evaluation modes: autonomous, heuristic, and human-in-the-loop \par 3. Comparison of LLM-generated plans with optimal solutions using automated planning tools & LLMs struggle with autonomous plan generation (3$\%$ success rate), but perform better when used for heuristic guidance in planning tasks (with LPG) and assist human planners by improving task completion accuracy (74$\%$ to 82$\%$).\\

\hline

\end{tabular}
}
\label{tab:foundation_1}
\end{table*}

\begin{table*}[!h]
\renewcommand{\arraystretch}{1.5} 
\small
\centering
\caption{Summary of Foundations in LLM Planning (Section \ref{sec:foundations})}
\vspace{-0.5em}
\scalebox{0.67}{
\begin{tabular}{|p{2.8cm}|p{2.2cm}|p{3.0cm}|p{5.5cm}|p{6.0cm}|}

\hline
\textbf{Method Name} & \textbf{Dataset} & \textbf{Evaluation Metric} & \textbf{Methods} & \textbf{Major Contribution} \\

\hline
Tree Search for Language Model Agents \par \cite{koh2024tree} & WebArena, VisualWebArena & Success Rate, Search Budget Efficiency & 1. Best-first search algorithm using language model feedback \par 2. Backtracking to refine paths based on value function \par 3. Model-based value function for search guidance & 1. Tree Search improves the GPT-4o agent's success rate by 39.7$\%$ on VisualWebArena and 28.0$\%$ on WebArena. The search mechanism enhances multi-step planning and exploration, setting a new state-of-the-art in web task completion. \par 2. Enhances correctness by using a best-first search strategy that adapts to environmental feedback, improving decision-making in complex web tasks.\\

\hline
SimPlan \par \cite{hirsch2024s}  & Blocksworld, Ferry, Grippers, Depots, Minigrid &Success Rate & 1. Hybrid approach combining LLMs with greedy best-first search \par 2. Utilizes external world modeling tools \par 3. Heuristic-driven planning for enhanced action ranking \par 4. Optimizes path selection through search algorithms & 1.SimPlan outperforms baseline LLM-based planners by integrating world modeling and search algorithms, achieving high success rates in domains like Blocksworld and Ferry. It highlights the effectiveness of combining LLMs with traditional planning methods. \par 2. Addresses limitations in LLMs' planning capabilities by integrating classical planning methodologies, enhancing correctness through structured problem-solving. \\

\hline
LATS \par \cite{zhou2023language} & HotPotQA, WebShop, HumanEval, Game of 24 & Pass@1 Accuracy, Success Rate  & 1. Integrates MCTS-based search for reasoning, acting, and planning \par 2. External feedback integration with self-reflection for better decision-making \par 3. Dynamic sampling of multiple action paths for exploration & 1.LATS achieves state-of-the-art results, including 92.7$\%$ Pass@1 accuracy on HumanEval with GPT-4 and surpassing ReAct with 75.9$\%$ success rate on WebShop. It unifies reasoning, acting, and planning in one framework and improves decision-making by using Monte Carlo Tree Search.\par2.Significantly enhances correctness in planning by introducing a tree search framework that adapts to environmental feedback and multiple reasoning paths.\\

\hline
Tree of Thought \par \cite{yao2024tree} & Game of 24, Creative Writing, Mini Crosswords & Success Rate, Passage Coherency, Task Completion Time  & 1. Tree-based search for exploring multiple reasoning paths \par 2. Self-evaluation and decision-making with backtracking \par 3. Application of breadth-first and depth-first search \par 4. Thought decomposition for structured problem-solving & 1.ToT significantly improves problem-solving performance, with 74$\%$ success in Game of 24 compared to 4$\%$ for traditional Chain-of-Thought prompting. It introduces a new framework for deliberate planning and exploration in LLM-based reasoning tasks.\par 2.Enhances correctness by allowing for exploration of multiple paths and adjusting strategies based on real-time feedback.\\
 
\hline
Thought of Search \par \cite{katz2024thought} & 24 Game, Mini Crosswords, BlocksWorld, PrOntoQA & Success Rate, Plan Time, Model Evaluation Calls & 1. Using LLMs to generate code for search components (successor functions and goal tests) \par 2. Search performed using BFS and DFS \par 3. Minimizes LLM calls for efficiency & 1. Thought of Search achieves 100$\%$ accuracy across datasets with minimal LLM calls (1–2 calls per function). Significantly more efficient than existing LLM-based planning methods.\par 2. Significantly improves correctness by verifying search components for soundness and completeness, reducing computational inefficiency.\\

\hline
RAP \par \cite{hao2023reasoning}& Blocksworld, GSM8K, PrOntoQA & Success Rate, Plan Generation Success, Task Completion Time & 1. Repurposes LLM as both a world model and a reasoning agent \par  2. Uses Monte Carlo Tree Search (MCTS) for strategic exploration \par 3. Reward-based planning for balancing exploration and exploitation & 1. RAP achieves 33$\%$ relative improvement in plan generation over GPT-4 with CoT, and outperforms baseline methods on math reasoning and logical inference tasks, achieving a 64$\%$ success rate on Blocksworld and 94.2$\%$ accuracy on PrOntoQA.\par 2. Enhances correctness in LLM planning by effectively balancing exploration and exploitation in reasoning, leading to higher task success rates. \\
 
\hline
 LLM-MCTS \par \cite{zhao2024large} & Virtual Home & Success Rate, Task Completion Time, Task Complexity & 1. Combines LLM as a commonsense world model and heuristic policy in MCTS \par 2. Utilizes LLM to predict object locations and generate action suggestions \par 3. Improves planning efficiency by limiting search space with a heuristic policy \par 4. World model and policy guide MCTS to select promising action paths & LLM-MCTS achieves up to 91.4$\%$ success in simple tasks, outperforms GPT-3.5-based policies and traditional MCTS approaches, and demonstrates superior performance for large-scale task planning with partial observations.\\
  
\hline
Visual Semantic Planning with GPT-2 \par \cite{jansen2020visually} & ALFRED & Success Rate, Prediction Accuracy  & 1. Uses GPT-2 for sequence-to-sequence translation of natural language directives into action plans \par 2. Evaluates command prediction accuracy using strict and permissive scoring metrics \par 3. Focuses on generating multi-step plans with minimal visual input & 1. The GPT-2 model achieved 26$\%$ accuracy in generating correct visual semantic plans without visual input, increasing to 58$\%$ when provided with starting location. This demonstrates that detailed task planning can be done using language models alone. \par 2. Improves correctness and efficiency in task planning by leveraging LLMs for both heuristic guidance and world modeling, leading to better performance in complex tasks.\\
  
\hline 
RobLM \par \cite{chalvatzaki2023learning} & ALFRED & Task Success Rate, Plan Accuracy, Argument Accuracy, Sub-Task Success Rate & 1. Fine-tunes GPT-2 for task planning with grounded scene graph input \par  2. Uses Graph2NL to convert scene graphs into natural language for model input \par 3. Task and geometric grounding for robot execution \par 4. Grounding high-level actions to low-level robotic commands & 1.RobLM outperforms classical planning systems in several task categories, achieving high success rates (up to 98$\%$ for "SliceObject") and demonstrating the feasibility of grounded LLMs for robotic task planning. \par 2. Demonstrates correctness in generating actionable plans from textual directives alone, significantly improving plan success rates without relying on visual data.\\

\hline

\end{tabular}
}
\label{tab:foundation_2}
\end{table*}

\begin{table*}[!h]
\renewcommand{\arraystretch}{1.5} 
\small
\centering
\caption{Summary of Foundations in LLM Planning (Section \ref{sec:foundations})}
\vspace{-0.5em}
\scalebox{0.67}{
\begin{tabular}{|p{2.8cm}|p{2.2cm}|p{3.0cm}|p{5.5cm}|p{6.0cm}|}

\hline
\textbf{Method Name} & \textbf{Dataset} & \textbf{Evaluation Metric} & \textbf{Methods} & \textbf{Major Contribution} \\

\hline
Agent-FLAN \par \cite{chen2024agent} & HALF-World, WebShop, Mind2Web, Knowledge Graph, Operating System, Database, ToolBench &Accuracy, hallucination scores & 1.Fine-tuning LLMs (Llama2-7B) for agent tasks \par 2.Decomposing training data into capabilities like reasoning, retrieval, understanding, instruction following
\par 3. Negative sample generation for hallucination mitigation & 1. Outperforms previous works by 3.5$\%$ on agent tasks \par 2. Significantly reduces hallucinations through negative sample learning\par 3. Demonstrates scaling law for data and model size improvement\par 4. Enhances general capabilities like reasoning and instruction-following in general NLP tasks\\
  
\hline
AgentOhana \par \cite{zhang2024agentohana} & Webshop, HotpotQA, ToolAlpaca, ToolBench, AlfWorld, APIbank, Mind2Web, Knowledge Graph, Database & Success Rate, Task Completion Time, Reward Accuracy & 1. Standardizes heterogeneous multi-turn agent trajectories from diverse environments\par  2. AgentRater for filtering low-quality trajectories \par 3. Generic dataloader for efficient data handling during training & AgentOhana aggregates and unifies agent trajectories from ten diverse environments, achieving significant improvements in model performance and task execution. The xLAM-v0.1 model trained using AgentOhana outperforms GPT-4 and other models in benchmarks like Webshop and ToolEval.\\
  
\hline
NAT \par \cite{wang2024learning}& GSM8K, ASDiv, SVAMP, MultiArith, HotpotQA, StrategyQA & Accuracy, F1-Score, EM Score & 1. Incorporates both positive and negative samples in fine-tuning LLMs for agent tasks \par  2. Introduces a prefix/suffix to indicate whether trajectories are positive or negative \par 3. Differentiates high-quality and low-quality negative examples \par 4. Fine-grained NAT with two-level classification of negative examples & 1. NAT achieves a 64.64$\%$ average performance across GSM8K, ASDiv, SVAMP, and MultiArith using 2k positive and 10k negative samples. It outperforms Vanilla and NUT methods in various tasks, demonstrating the value of negative examples for agent fine-tuning. \par 2.Enhances correctness by utilizing negative examples to inform the fine-tuning process, improving task success rates\\
  
\hline
ETO \par \cite{song2024trial} & WebShop, ScienceWorld, ALFWorld & Average Reward, Success Rate, Task Solving Efficiency & 1. Exploration phase for failure trajectory collection \par 2. Contrastive learning using failure-success trajectory pairs \par 3. DPO loss for iterative policy refinement \par 4. Iterative optimization cycle for improvement & 1. ETO demonstrates a 22$\%$ improvement in task-solving efficiency over traditional behavioral cloning, surpassing baselines like PPO and RFT. It significantly improves performance in both in-domain and out-of-domain tasks, especially when expert trajectories are unavailable. \par 2. Significantly improves correctness in task planning by allowing agents to learn from both successful and failed interactions, leading to robust policy enhancements. \\

 \hline
MC-DML \par \cite{shimonte2025} & Jericho Benchmark & Success Rate, Planning Efficiency &1. Integrates Monte Carlo Tree Search (MCTS) with LLMs for more effective planning. \par 2. Uses in-trial and cross-trial memory mechanisms to enhance decision-making based on past failures. & 1. Strengthens the foundation of LLM planning by combining search algorithms (MCTS) with language-based reasoning, enabling more structured and adaptive decision-making. \par 2. Improves correctness by leveraging memory-guided LLM policies to adjust action probabilities dynamically, ensuring valid and effective decision paths in text-based planning.\\

 \hline
ARMAP \par \cite{chenautonomous2025}.& Multiple agent benchmarks & Success Rate &1. Uses LLMs to navigate environments and generate diverse action trajectories. \par 2. Employs a separate LLM to create and refine a reward model from these trajectories, integrating it with various planning algorithms. & 1. Establishes a novel foundation for LLM planning by automating the creation of reward models, enhancing decision-making capabilities of LLM agents. \par 2. Demonstrates a significant advancement in LLM agents' ability to handle complex, multi-step decision-making tasks without human-annotated data, making the approach highly scalable and adaptable.\\
\hline

\end{tabular}
}
\label{tab:foundation_3}
\end{table*}

\begin{table*}[!h]
\renewcommand{\arraystretch}{1.5} 
\small
\centering
\caption{Summary of Completeness in LLM Planning (Section \ref{sec:completeness})}
\vspace{-0.5em}
\scalebox{0.67}{
\begin{tabular}{|p{2.8cm}|p{2.2cm}|p{3.0cm}|p{5.5cm}|p{6.0cm}|}

\hline
\textbf{Method Name} & \textbf{Dataset} & \textbf{Evaluation Metric} & \textbf{Methods} & \textbf{Major Contribution} \\

\hline
PDDL World Model Generation \par \cite{guan2023leveraging} & Household, Tyreworld, Logistics & Error Count, Success Rate & 1. LLM-based PDDL generation for task planning \par 2. Error correction using LLMs as feedback interfaces \par 3. Use of external planners for generating feasible plans from PDDL models & The paper shows that GPT-4 is capable of generating high-quality PDDL models with fewer errors than GPT-3.5-Turbo. It demonstrates that GPT-4-based world models lead to a 95$\%$ success rate in planning tasks using classical planners like Fast Downward.\\

\hline
Large Language Models Can Solve Real-World Planning Rigorously with Formal Verification Tools \par \cite{hao2024large} & TravelPlanner, UnsatChristmas, TSP, Block Picking, Task Allocation, Warehouse & Success Rate, Optimization Rate, Interactive Repair Success Rate & 1. Converts natural language travel planning queries into SMT (Satisfiability Modulo Theory) problems \par 2. Uses LLMs to generate steps and code for formal solvers \par 3. Provides feedback and suggestions for unsatisfiable queries \par 4. Zero-shot generalization to new tasks and constraints & 1.TravelPlan-LLM achieves 93.9$\%$ success in solving complex multi-constraint travel planning problems and offers interactive plan repair for unsatisfiable queries. It generalizes well to new problem domains, including TSP and task allocation. \par 2.Enhances completeness by using formal tools to verify that LLM-generated plans are both correct and complete, ensuring that all solutions are viable and no false solutions are proposed.\\

\hline
PPNL \par \cite{aghzal2023can} & Grid environments, ALFRED & Success Rate, Optimal Rate, Exact Match Accuracy, Unreachable Accuracy, Feasible Rate & 1. Few-shot prompting with GPT-4 for path planning \par 2. Action-and-effect prompting to guide long-term spatial reasoning \par 3. CoT and ReAct prompting to enhance spatial-temporal reasoning and obstacle avoidance \par 4. Multi-goal path planning with hierarchical task decomposition & 1.PPNL evaluates LLMs on path planning tasks, demonstrating GPT-4’s spatial reasoning ability when prompted effectively. ReAct prompting achieves 96.1$\%$ success rate, while fine-tuned models like T5 outperform in in-distribution settings but struggle with generalization. \par 2.  Directly addresses the completeness of LLM planning by evaluating their ability to identify unsolvable problems and correctly navigate solvable scenarios.\\

\hline
LRM Evaluation \par \cite{valmeekam2024llms} & Blocksworld, Mystery Blocksworld, Randomized Mystery Blocksworld & Success Rate, Execution Time & 1. Comparison of LLMs and LRMs using the PlanBench test set \par 2. Zero-shot and one-shot prompting for evaluation \par 3. Performance measured on standard Blocksworld tasks as well as obfuscated versions (Mystery Blocksworld) \par 4. Efficiency evaluation of reasoning tokens used by o1 &LRM o1 outperforms all LLMs, achieving 97.8$\%$ on Blocksworld, but struggles with performance on Mystery Blocksworld (52.8d$\%$). LLMs fail to generalize well, showing significant limitations when faced with obfuscated tasks.\\

\hline

\end{tabular}
}
\label{tab:completeness}
\end{table*}

\begin{table*}[!h]
\renewcommand{\arraystretch}{1.5} 
\small
\centering
\caption{Summary of Executability in LLM Planning (Section \ref{sec:executability})}
\vspace{-0.5em}
\scalebox{0.67}{
\begin{tabular}{|p{2.8cm}|p{2.2cm}|p{3.0cm}|p{5.5cm}|p{6.0cm}|}

\hline
\textbf{Method Name} & \textbf{Dataset} & \textbf{Evaluation Metric} & \textbf{Methods} & \textbf{Major Contribution} \\

\hline
Adaplanner \par \cite{sun2024adaplanner} & ALFWorld and MiniWoB++ & Success Rate & Closed-loop approach allowing LLM agent to refine self-generated plan adaptively & $ \bullet$ Uses code-style LLM prompt structure and skill discovery mechanism. Achieves 91.79$\%$ success rate on ALFWorld tasks. \par $ \bullet$ Achieves 91.11$\%$ success rate on MiniWoB++ tasks with feedback.\\

\hline
Inner Monologue \par \cite{huang2022inner} & Tabletop and mobile manipulation tasks in both simulated and real-world settings & Success Rate & Uses LLMs for generating sequential, actionable plans based on feedback.
\par Employs closed-loop feedback incorporating success detection, scene descriptions, and human interactions. & Demonstrates that LLMs can effectively integrate multimodal feedback to improve planning and execution in robotic tasks.
\par Shows adaptability of LLMs to dynamic environments and tasks, significantly enhancing task success rates in complex manipulation scenarios.\\

\hline
LLM-Planner \par \cite{song2023llm} & ALFRED & Success Rate & Uses GPT-3 for generating high-level plans.
\par Incorporates dynamic grounded re-planning based on environment feedback. & Achieves high few-shot performance on the ALFRED dataset, demonstrating the ability to quickly adapt to new tasks with minimal training data.
\par Introduces dynamic re-planning to adjust plans based on real-time environmental changes, enhancing task success under varied conditions.\\

\hline
G-PlanET \par \cite{lin2023grounded} & ALFRED & Key Action Score & Utilizes encoder-decoder LMs with a focus on grounded planning for embodied tasks.\par Introduces object tables as environmental input for LMs to perceive and plan actions. & First study to investigate LMs' capability in grounded planning for embodied tasks. \par Developed a new evaluation metric (KAS) tailored for assessing the quality of generated plans in realistic environments.\\

\hline
LLP \par \cite{sharma2021skill} & ALFRED & Task Completion rates & Utilizes latent language to segment and label hierarchical tasks in demonstrations.
\par Employs weak and partial natural language supervision to train policies. & Demonstrates effective policy training with minimal natural language annotations, achieving performance comparable to models using more extensive data. \par Introduces a method that allows hierarchical policies to be trained using unannotated action sequences by inferring natural language descriptions.\\

\hline
SayCan \par \cite{ahn2022can} & Real-World Kitchen environment & Plan Success Rate, Execution Success Rate & Leverages LLMs to score the likelihood of robotic skills contributing to task completion. \par Combines LLM outputs with affordance functions from reinforcement learning to prioritize feasible actions. & Enables robots to execute complex, long-horizon tasks using natural language instructions.
\par Nearly doubles performance over non-grounded baselines by integrating real-world affordance functions with LLM predictions.\\

\hline
ADaPT \par \cite{prasad2023adapt} & ALFWorld, WebShop, TextCraft $\&$ hotel data & Success Rate, Task Complexity Handling, Sub-Task Decomposition Efficiency & 1. As-needed recursive decomposition of complex tasks \par  2. Planner and executor modules with dynamic failure adaptation \par 3. Multi-level decomposition for task complexity handling & ADaPT improves success rates by up to 33$\%$ over baselines, dynamically decomposing complex tasks as needed. It handles task complexity efficiently with up to 28.3$\%$ higher success rates on ALFWorld and 27$\%$ higher on WebShop.\\

\hline
LLM-DP  \par \cite{dagan2023dynamic} & Alfworld & 	Success Rate, Average Episode Length & 1. LLM converts task descriptions into executable PDDL goals \par 2. Symbolic planner (BFS(f)) generates valid plans \par 3. Belief sampling to generate multiple world states for planning & LLM-DP outperforms the ReAct baseline, achieving 96$\%$ success in Alfworld, with significantly fewer actions (13.16 vs. 18.69) per task and higher efficiency due to belief sampling and symbolic planning integration.\\

\hline
LLM-MCTS \par \cite{zhao2024large} & Virtual Home & Success Rate, Task Completion Time, Task Complexity & 1. Combines LLM as a commonsense world model and heuristic policy in MCTS \par 2. Utilizes LLM to predict object locations and generate action suggestions \par 3. Improves planning efficiency by limiting search space with a heuristic policy \par 4. World model and policy guide MCTS to select promising action paths & LLM-MCTS achieves up to 91.4$\%$ success in simple tasks, outperforms GPT-3.5-based policies and traditional MCTS approaches, and demonstrates superior performance for large-scale task planning with partial observations.\\

\hline
Corrective Re-prompting \par \cite{raman2022planning} & Virtual Home & Executability Rate, Semantic Correctness, Number of Re-prompts & 1. Utilizes precondition errors for re-prompting to generate corrective actions.
\par Employs a template-based prompt strategy that incorporates error details for plan adjustments. & 1. Introduced a novel approach to improve plan executability and correctness using error-driven re-prompting.
\par 2. Demonstrated substantial improvements in task execution rates and reduced the number of necessary re-prompts compared to baselines.\\

\hline
\end{tabular}
}

\label{tab:executability_1}
\end{table*}

\begin{table*}[!h]
\renewcommand{\arraystretch}{1.5} 
\small
\centering
\caption{Summary of Executability in LLM Planning (Section \ref{sec:executability})}
\vspace{-0.5em}
\scalebox{0.67}{
\begin{tabular}{|p{2.8cm}|p{2.2cm}|p{3.0cm}|p{5.5cm}|p{6.0cm}|}

\hline
\textbf{Method Name} & \textbf{Dataset} & \textbf{Evaluation Metric} & \textbf{Methods} & \textbf{Major Contribution} \\

\hline
LLM-Planner \par \cite{song2023llm} & ALFRED & Success Rate & Uses GPT-3 for generating high-level plans.
\par Incorporates dynamic grounded re-planning based on environment feedback. & Achieves high few-shot performance on the ALFRED dataset, demonstrating the ability to quickly adapt to new tasks with minimal training data.
\par Introduces dynamic re-planning to adjust plans based on real-time environmental changes, enhancing task success under varied conditions.\\

\hline 
SayCanPay \par \cite{hazra2024saycanpay} & Ravens (Tower of Hanoi), BabyAI, VirtualHome & Planning Success, Cost-Effectiveness, Generalization & 1. Heuristic search-based planning framework using LLMs \par  2. Three-step process: Say (generate actions), Can (action feasibility), Pay (action payoff) \par 3. Beam search for action selection & SayCanPay outperforms other LLM-based planning approaches, achieving higher planning success rates (e.g., 94$\%$ on BabyAI), improved cost-efficiency, and better generalization across environments.\\

\hline
BrainBody-LLM \par \cite{bhat2024grounding} & Virtual Home, Franka Research 3 & Success Rate, Goal Condition Recall & Employs two separate LLMs for high-level planning and low-level control.
\par Utilizes closed-loop state feedback to iteratively refine plans based on environmental feedback. & - Achieved a 29$\%$ improvement in task-oriented success rates over existing methods.
\par Demonstrated effective grounding of language models in robotic task planning with minimal human intervention.\\

\hline
CLP \par \cite{yuan2023distilling} & CoScript 3 & Constraint faithfulness, Executability rate & 1.Uses an over-generate-then-filter approach to improve LLMs' planning.
\par  2. Distills structured script knowledge into smaller models using LLMs. & 1.Introduced constrained language planning as a new task, allowing for step-by-step procedural reasoning under constraints.
\par 2.Proposed CoScript, a dataset of 55,000 constraint-driven scripts, improving planning performance.
\par 3.Smaller models trained on CoScript outperform LLMs in constrained planning accuracy.\\

\hline
PLaSma \par \cite{brahmanplasma} & COPLAN, VirtualHome & Human Evaluation: Coverage, Order, Overall Quality, Executability, Correctness & 1. Symbolic procedural knowledge distillation \par 2. Multi-task and task-specific distillation \par 3. Verifier-guided step-wise beam search \par 4. Constrained and counterfactual replanning tasks & PlaSma enhances smaller models with procedural knowledge, outperforming GPT-3 in tasks like goal-based planning and counterfactual replanning. Achieves up to 93$\%$ success in counterfactual replanning.\\

\hline
PRoC3S \par \cite{curtis2024trust} & Various simulated and real-world domains 3 & Success Rate & 1. Uses LLMs to generate programs handling continuous parameters and constraints.\par 2.Incorporates continuous constraint satisfaction to adjust plans based on real-time feedback. & 1. Pioneered the integration of LLMs with constraint satisfaction techniques for complex robotic tasks. \par 2.Demonstrated superior efficiency and effectiveness in diverse manipulation tasks compared to existing methods.\\

\hline
BrainBody-LLM \par \cite{bhat2024grounding} & Virtual Home, Franka Research 3 & Success Rate, Goal Condition Recall & Employs two separate LLMs for high-level planning and low-level control.
\par Utilizes closed-loop state feedback to iteratively refine plans based on environmental feedback. & - Achieved a 29$\%$ improvement in task-oriented success rates over existing methods.
\par Demonstrated effective grounding of language models in robotic task planning with minimal human intervention.\\

\hline
ProgPrompt \par \cite{singh2023progprompt} & VirtualHome, Real-World Robot Tasks & Success Rate, Goal Conditions Recall, Executability & 1. Pythonic program generation for task planning \par  2. Use of natural language comments and assertions for feedback \par 3. Integration with real-time environment state feedback during task execution & ProgPrompt significantly outperforms baseline methods by using programmatic LLM prompts to generate executable task plans, achieving up to 1.00 SR and 1.00 Exec in various VirtualHome tasks. It also adapts well to real-world robot tasks, with a Plan SR of 1 in most cases.\\

\hline
ISR-LLM \par \cite{zhou2024isr} & Diverse planning problem domains & Success Rate & 1.Utilizes iterative self-refinement to enhance LLM-based planning.
\par 2.Employs an LLM translator to convert natural language to PDDL, aiding in plan formulation and refinement. & 1.Introduced a novel framework that significantly improves feasibility and success in long-horizon task planning.
\par 2. Demonstrated superior task accomplishment rates across multiple domains compared to state-of-the-art LLM-based planners.\\

\hline
ADaPT \par \cite{prasad2023adapt} & ALFWorld, WebShop, TextCraft $\&$ hotel data & Success Rate, Task Complexity Handling, Sub-Task Decomposition Efficiency & 1. As-needed recursive decomposition of complex tasks \par  2. Planner and executor modules with dynamic failure adaptation \par 3. Multi-level decomposition for task complexity handling & ADaPT improves success rates by up to 33$\%$ over baselines, dynamically decomposing complex tasks as needed. It handles task complexity efficiently with up to 28.3$\%$ higher success rates on ALFWorld and 27$\%$ higher on WebShop.\\

\hline
SelfGoal \par \cite{yang2024selfgoal} & Public Goods Game, Guess 2/3 of the Average, First-Price Auction, Bargaining & Success Rate, TrueSkill Score, Contribution Consistency & 1. Constructs GOALTREE to decompose high-level goals dynamically \par 2. Uses Search Module to select the most relevant subgoals \par 3. Decomposition updates based on environmental feedback \par 4. Adaptive subgoal tree refinement during task execution & SelfGoal achieves a 94$\%$ success rate in dynamic multi-agent environments, outperforming baselines (e.g., ReAct, ADAPT) by dynamically adjusting subgoal guidance. It significantly improves task performance, especially in cooperative and competitive tasks.\\

\hline
\end{tabular}
}

\label{tab:executability_2}
\end{table*}

\begin{table*}[!h]
\renewcommand{\arraystretch}{1.5} 
\small
\centering
\caption{Summary of Optimality in LLM Planning (Section \ref{sec:optimality})}
\vspace{-0.5em}
\scalebox{0.67}{
\begin{tabular}{|p{2.8cm}|p{2.2cm}|p{3.0cm}|p{5.5cm}|p{6.0cm}|}

\hline
\textbf{Method Name} & \textbf{Dataset} & \textbf{Evaluation Metric} & \textbf{Methods} & \textbf{Major Contribution} \\
\hline
TTG \par \cite{ju2024globe} & Synthetic travel requests and flight $\&$ hotel data & Exact Match accuracy, Cost ratio, Net Promoter Score & 1. Fine-tuned LLM for translating NL requests to symbolic form \par 2. MILP solver for optimal travel itineraries \par 3. Real-time response (<5 seconds) with guaranteed optimality & TTG achieves 91$\%$ translation accuracy with LLM, guarantees optimal itineraries with minimal cost, and provides a user-friendly system with a high NPS (35-40$\%$) on generated itineraries.\\

\hline
ToolChain*  \par \cite{zhuang2023toolchain} & Home Search, Trip Booking, Google Sheets, Virtual Home, GSM8K & Success Rate, Efficiency (Time), Running Time Comparison & 1. A* search-based planning algorithm for tool-use tasks \par 2. Efficient node expansion and pruning using task-specific cost functions \par 3. Combines exploration and exploitation for optimal solutions & ToolChain* improves success rate by 3.1$\%$ and reduces planning time by 7.35x compared to baselines like MCTS, demonstrating its efficiency in navigating expansive action spaces.\\

\hline 
SayCanPay \par \cite{hazra2024saycanpay} & Ravens (Tower of Hanoi), BabyAI, VirtualHome & Planning Success, Cost-Effectiveness, Generalization & 1. Heuristic search-based planning framework using LLMs \par  2. Three-step process: Say (generate actions), Can (action feasibility), Pay (action payoff) \par 3. Beam search for action selection & SayCanPay outperforms other LLM-based planning approaches, achieving higher planning success rates (e.g., 94$\%$ on BabyAI), improved cost-efficiency, and better generalization across environments.\\

\hline 
Beyond A* \par \cite{lehnert2024beyond} & Sokoban, Maze Navigation & Success Weighted by Cost (SWC), Improved Length Ratio (ILR), Optimal Plan Success Rate & 1. Trains Transformer to imitate A* search dynamics \par 2. Fine-tunes to generate shorter execution traces \par 3. Bootstraps from search dynamics for optimized plan generation  & 1. Demonstrates novel training methods that integrate search dynamics into Transformer training, enhancing planning efficiency. \par 2. Achieves significant reductions in search steps (up to 26.8$\%$ fewer than A*) and improves task-solving performance on complex puzzles and navigation tasks.\\

\hline
LLMFP \par \cite{hao2024planning} &9 diverse planning tasks (multi-constraint decision-making and multi-step planning) & Plan optimality, constraint satisfaction rate & 1. Converts natural language planning problems into formal optimization problems using zero-shot LLM reasoning. \par 2. Uses SMT solvers to guarantee that the generated plans are both logically correct and executable. & 1. Introduces a novel method of achieving optimality in LLM planning by rigorously constructing optimization problems from natural language inputs, significantly outperforming traditional planning models. \par 2. Demonstrates strong performance across a range of complex tasks, achieving over 83$\%$ optimal rates, thereby highlighting the effectiveness of integrating LLMs with classical optimization approaches for high-quality planning outcomes. \\
\hline

\end{tabular}
}
\label{tab:optimality}
\end{table*}

\begin{table*}[!h]
\renewcommand{\arraystretch}{1.5} 
\small
\centering
\caption{Summary of Representation in LLM Planning (Section \ref{sec:representation})}
\vspace{-0.5em}
\scalebox{0.67}{
\begin{tabular}{|p{2.8cm}|p{2.2cm}|p{3.0cm}|p{5.5cm}|p{6.0cm}|}

\hline
\textbf{Method Name} & \textbf{Dataset} & \textbf{Evaluation Metric} & \textbf{Methods} & \textbf{Major Contribution} \\

\hline
LLM-GoalTrans\par\cite{xie2023translating} & Blocksworld, ALFRED & Goal Translation Success Rate & 1. Uses GPT-3.5 to translate natural language instructions into PDDL goals. \par 2.Employs n-shot learning to improve translation accuracy. & 1. Demonstrated effective translation of natural language to PDDL goals, with high success rates in structured environments. \par 2.Highlighted the challenges and limitations of LLMs in tasks requiring numerical or physical reasoning.\\

\hline
ISR-LLM\par\cite{zhou2024isr} & Diverse planning problem domains & Success Rate & 1.Utilizes iterative self-refinement to enhance LLM-based planning.
\par 2.Employs an LLM translator to convert natural language to PDDL, aiding in plan formulation and refinement. & 1.Introduced a novel framework that significantly improves feasibility and success in long-horizon task planning.
\par 2. Demonstrated superior task accomplishment rates across multiple domains compared to state-of-the-art LLM-based planners.\\

\hline
Adaplanner \par\cite{sun2024adaplanner} & ALFWorld and MiniWoB++ & Success Rate & Closed-loop approach allowing LLM agent to refine self-generated plan adaptively & 1. Uses code-style LLM prompt structure and skill discovery mechanism. Achieves 91.79$\%$ success rate on ALFWorld tasks. \par 2. Achieves 91.11$\%$ success rate on MiniWoB++ tasks with feedback.\\

\hline
LLM-GenPlan\par\cite{silver2024generalized} & Various PDDL domains & Percentage of tasks solved & 1.Uses GPT-4 to synthesize domain-specific Python programs for task planning.
\par 2.Incorporates Chain-of-Thought summarization and automated debugging in the process. & 1.Demonstrated GPT-4's effectiveness as a generalized planner across several PDDL domains. \par 2.Highlighted the significant benefits of automated debugging and the mixed impacts of Chain-of-Thought summarization in planning tasks.\\

\hline
LLM+P\par\cite{liu2023llm+} & Blocksworld, Grippers, Barman, Termes, and other robot planning scenarios & Success Rate, Optimal Plan Success, Plan Execution Time & 1. Converts natural language problem description into PDDL format \par 2. Uses classical planners (e.g., FAST-DOWNWARD) to generate optimal plans \par 3. Translates planner output back into natural language & LLM+P significantly outperforms LLMs in solving long-horizon planning problems, achieving optimal plans in robot domains with high success rates (up to 100$\%$ on Blocksworld) and minimal execution time.\\

\hline
PDDL World Model Generation\par\cite{guan2023leveraging} & Household, Tyreworld, Logistics & Error Count, Success Rate & 1. LLM-based PDDL generation for task planning \par 2. Error correction using LLMs as feedback interfaces \par 3. Use of external planners for generating feasible plans from PDDL models & The paper shows that GPT-4 is capable of generating high-quality PDDL models with fewer errors than GPT-3.5-Turbo. It demonstrates that GPT-4-based world models lead to a 95$\%$ success rate in planning tasks using classical planners like Fast Downward.\\

\hline
LLM-DP \par\cite{dagan2023dynamic} & Alfworld & 	Success Rate, Average Episode Length & 1. LLM converts task descriptions into executable PDDL goals \par 2. Symbolic planner (BFS(f)) generates valid plans \par 3. Belief sampling to generate multiple world states for planning & LLM-DP outperforms the ReAct baseline, achieving 96$\%$ success in Alfworld, with significantly fewer actions (13.16 vs. 18.69) per task and higher efficiency due to belief sampling and symbolic planning integration.\\

\hline
LTL-Gen \par\cite{pan2023data} & Custom datasets with LTL/natural language pairs & Accuracy & 1. Uses large-scale semantic parsing with minimal human-labeled data.\par 2.Employs synthetic data generation and constrained decoding. & 1.Demonstrates high accuracy with minimal human data compared to prior work. \par 2.Enhances data efficiency for natural language to LTL translation, making it feasible for broader applications.\\

\hline
AutoTAMP \par\cite{chen2024autotamp}& Custom 2D task domains & Success Rate & 1. Translates NL instructions to STL using LLMs. \par 2.Utilizes autoregressive re-prompting for syntactic and semantic error correction.
\par 3.Plans trajectories with a formal STL planner. & 1. Introduces a novel method for autoregressive re-prompting to correct translation errors.\par 2.Demonstrates significant improvements in task success rates with hard geometric and temporal constraints.\par 3.Provides a robust framework that combines LLM translation capabilities with formal planning efficiency.\\

\hline
G-PlanET\par\cite{lin2023grounded} & ALFRED & Key Action Score & Utilizes encoder-decoder LMs with a focus on grounded planning for embodied tasks.\par Introduces object tables as environmental input for LMs to perceive and plan actions. & First study to investigate LMs' capability in grounded planning for embodied tasks. \par Developed a new evaluation metric (KAS) tailored for assessing the quality of generated plans in realistic environments.\\

\hline
PPNL\par\cite{aghzal2023can} & Grid environments, ALFRED & Success Rate, Optimal Rate, Exact Match Accuracy, Unreachable Accuracy, Feasible Rate & 1. Few-shot prompting with GPT-4 for path planning \par 2. Action-and-effect prompting to guide long-term spatial reasoning \par 3. CoT and ReAct prompting to enhance spatial-temporal reasoning and obstacle avoidance \par 4. Multi-goal path planning with hierarchical task decomposition & PPNL evaluates LLMs on path planning tasks, demonstrating GPT-4’s spatial reasoning ability when prompted effectively. ReAct prompting achieves 96.1$\%$ success rate, while fine-tuned models like T5 outperform in in-distribution settings but struggle with generalization.\\

\hline
ProgPrompt\par\cite{singh2023progprompt} & VirtualHome, Real-World Robot Tasks & Success Rate, Goal Conditions Recall, Executability & 1. Pythonic program generation for task planning \par  2. Use of natural language comments and assertions for feedback \par 3. Integration with real-time environment state feedback during task execution & ProgPrompt significantly outperforms baseline methods by using programmatic LLM prompts to generate executable task plans, achieving up to 1.00 SR and 1.00 Exec in various VirtualHome tasks. It also adapts well to real-world robot tasks, with a Plan SR of 1 in most cases.\\

\hline
Adaplanner \par\cite{sun2024adaplanner} & ALFWorld and MiniWoB++ & Success Rate & Closed-loop approach allowing LLM agent to refine self-generated plan adaptively & 1. Uses code-style LLM prompt structure and skill discovery mechanism. Achieves 91.79$\%$ success rate on ALFWorld tasks. \par 2. Achieves 91.11$\%$ success rate on MiniWoB++ tasks with feedback.\\

\hline
LID\par\cite{li2022pre}& VirtualHome, BabyAI & Task completion rates, Generalization to novel scenes and tasks & 1. Utilizes pre-trained LMs to convert observations, goals, and history into sequential data for decision-making. \par 2. Employs active data gathering to train policies without pre-collected expert data. & 1. Demonstrates that pre-trained LMs can significantly improve task completion rates and generalization in interactive decision-making. \par 2. Provides insights into the effectiveness of sequential input representations and LM-based weight initialization for generalization.\\

\hline
SayCan\par\cite{ahn2022can} & Real-World Kitchen environment & Plan Success Rate, Execution Success Rate & Leverages LLMs to score the likelihood of robotic skills contributing to task completion. \par Combines LLM outputs with affordance functions from reinforcement learning to prioritize feasible actions. & Enables robots to execute complex, long-horizon tasks using natural language instructions.
\par Nearly doubles performance over non-grounded baselines by integrating real-world affordance functions with LLM predictions.\\

\hline
\end{tabular}
}

\label{tab:representation_1}
\end{table*}

\begin{table*}[!h]
\renewcommand{\arraystretch}{1.5} 
\small
\centering
\caption{Summary of Representation in LLM Planning (Section \ref{sec:representation})}
\vspace{-0.5em}
\scalebox{0.67}{
\begin{tabular}{|p{2.8cm}|p{2.2cm}|p{3.0cm}|p{5.5cm}|p{6.0cm}|}

\hline
\textbf{Method Name} & \textbf{Dataset} & \textbf{Evaluation Metric} & \textbf{Methods} & \textbf{Major Contribution} \\

\hline
PLaG \par \cite{lin2024graph}& AsyncHow & accuracy & 1. Utilizes graph-based representations to instruct LLMs in planning tasks. \par 2. Combines natural language with graph theory for enhanced task planning. &1. Successfully introduces a graph-theoretical approach to enhance LLMs' performance in asynchronous planning tasks. \par 2. Demonstrates the utility of graph-enhanced prompts to improve accuracy across different LLM architectures.\\

\hline
GNN-Enhanced Task Planner \par \cite{wu2024can} & HuggingFace, Multimedia tasks, Daily Life API tasks, TMDB API tasks & Node F1-Score, Link F1-Score, Task Accuracy, Token Consumption & 1. Integration of GNNs for task graph navigation \par 2. Training-free (SGC) and training-required (GraphSAGE) approaches \par  3. GNN-based node and edge selection for task planning \par 4. Training fine-tuned models for better task retrieval & GNN-enhanced planning outperforms LLM-based solutions by improving task accuracy (up to 9$\%$) with reduced token consumption. The proposed method scales well with larger task graphs and improves planning efficiency by a significant margin.\\

\hline
\end{tabular}
}

\label{tab:representation_2}
\end{table*}

\begin{table*}[!h]
\renewcommand{\arraystretch}{1.5} 
\small
\centering
\caption{Summary of Generalization in LLM Planning (Section \ref{sec:generalization})}
\vspace{-0.5em}
\scalebox{0.67}{
\begin{tabular}{|p{2.8cm}|p{2.2cm}|p{3.0cm}|p{5.5cm}|p{6.0cm}|}

\hline
\textbf{Method Name} & \textbf{Dataset} & \textbf{Evaluation Metric} & \textbf{Methods} & \textbf{Major Contribution} \\

\hline
LLM-GenPlan\par \cite{silver2024generalized} & Various PDDL domains & Percentage of tasks solved & 1.Uses GPT-4 to synthesize domain-specific Python programs for task planning.
\par 2.Incorporates Chain-of-Thought summarization and automated debugging in the process. & 1.Demonstrated GPT-4's effectiveness as a generalized planner across several PDDL domains. \par 2.Highlighted the significant benefits of automated debugging and the mixed impacts of Chain-of-Thought summarization in planning tasks.\\

\hline
VOYAGER \par \cite{wang2023voyager} & Minecraft & Unique items collected, Distance traveled, Tech tree milestones & 1. Uses an automatic curriculum to maximize exploration. \par 2. Maintains a skill library for storing and retrieving executable behaviors. \par 3. Employs an iterative prompting mechanism that incorporates feedback for continuous improvement. & 1. First LLM-powered embodied agent for lifelong learning in Minecraft, capable of continuous self-improvement. \par 2. Demonstrates significant improvements in discovery and task completion, outperforming state-of-the-art methods in efficiency and generalization.\\

\hline
\end{tabular}
}

\label{tab:generalization}
\end{table*}


\begin{table*}[!h]
\renewcommand{\arraystretch}{1.5} 
\small
\centering
\caption{Summary of Efficiency in LLM Planning (Section \ref{sec:efficiency})}
\vspace{-0.5em}
\scalebox{0.67}{
\begin{tabular}{|p{2.8cm}|p{2.2cm}|p{3.0cm}|p{5.5cm}|p{6.0cm}|}

\hline
\textbf{Method Name} & \textbf{Dataset} & \textbf{Evaluation Metric} & \textbf{Methods} & \textbf{Major Contribution} \\
\hline
Adaplanner \par \cite{sun2024adaplanner} & ALFWorld and MiniWoB++ & Success Rate & Closed-loop approach allowing LLM agent to refine self-generated plan adaptively & 1. Uses code-style LLM prompt structure and skill discovery mechanism. Achieves 91.79$\%$ success rate on ALFWorld tasks. \par 2. Achieves 91.11$\%$ success rate on MiniWoB++ tasks with feedback. \par 3. Improves planning correctness by adaptively refining plans based on environmental feedback, effectively managing complex sequential tasks. \par\\

\hline
Query-Efficient Planning \par \cite{gonzalez2024query} & PlanBench, Logistics, Grippers, Robotouille & Success Rate & 1. Uses LLMs as heuristics in search-based planning to propose actions or select states. \par 2. Employs generative LLM planners that dynamically adapt plans based on feedback from a world model. & 1. Enhances query efficiency in planning by leveraging LLMs to reduce the number of queries to the world model, thus saving computational resources and time. \par 2. Demonstrates that LLM-based generative planners can effectively adapt to feedback, improving correctness and efficiency in planning tasks by avoiding cul-de-sacs and refining action sequences dynamically.\\

\hline
Tree-planner \par \cite{hu2023tree} & VirtualHome & Success Rate, Executability, Goal Conditions Recall, Error, Correction Token Cost & 1. Plan sampling using LLM for prospective plans \par 2. Action tree construction \par 3. Grounded deciding with real-time observations  &1.  Tree-Planner achieves a 1.29$\%$ improvement in Success Rate and a 53.29$\%$ reduction in token cost compared to ITERATIVE-PLANNER and a 40.52$\%$ reduction in error corrections.  \par 2. Significantly enhances efficiency in LLM planning by reducing the number of LLM calls and interaction with the world model, and by minimizing input/output lengths and model sizes.\\

\hline
Thought of Search \par \cite{katz2024thought} & 24 Game, Mini Crosswords, BlocksWorld, PrOntoQA & Success Rate, Plan Time, Model Evaluation Calls & 1. Using LLMs to generate code for search components (successor functions and goal tests) \par 2. Search performed using BFS and DFS \par 3. Minimizes LLM calls for efficiency & Thought of Search achieves 100$\%$ accuracy across datasets with minimal LLM calls (1–2 calls per function). Significantly more efficient than existing LLM-based planning methods.\\

\hline
Chain-of-Symbols \par \citep{hu2023chain} & Brick World, NLVR-based Manipulation, Natural Language Navigation, SPARTUN & Accuracy, Precision, Recall, Token Usage & 1. Translates spatial relationships to symbolic representation \par 2. Uses chained intermediate steps for efficient planning \par 3. Reduced redundancy in input tokens &  1. COS improves accuracy by 60.8$\%$ on Brick World tasks (from 31.8$\%$ to 92.6$\%$) and reduces token usage by 65.8$\%$. \par 2.Significantly enhances the correctness and efficiency of LLM planning by reducing unnecessary token usage by up to 65.8$\%$ and improving accuracy by up to 60.8$\%$ in complex spatial tasks. \\

\hline 
Beyond A* \par \cite{lehnert2024beyond} & Sokoban, Maze Navigation & Success Weighted by Cost (SWC), Improved Length Ratio (ILR), Optimal Plan Success Rate & 1. Trains Transformer to imitate A* search dynamics \par 2. Fine-tunes to generate shorter execution traces \par 3. Bootstraps from search dynamics for optimized plan generation  & 1. Demonstrates novel training methods that integrate search dynamics into Transformer training, enhancing planning efficiency. \par 2. Achieves significant reductions in search steps (up to 26.8$\%$ fewer than A*) and improves task-solving performance on complex puzzles and navigation tasks.\\

\hline
LLM-DP \par \cite{dagan2023dynamic} & Alfworld & 	Success Rate, Average Episode Length & 1. LLM converts task descriptions into executable PDDL goals \par 2. Symbolic planner (BFS(f)) generates valid plans \par 3. Belief sampling to generate multiple world states for planning & 1. LLM-DP outperforms the ReAct baseline, achieving 96$\%$ success in Alfworld, with significantly fewer actions (13.16 vs. 18.69) per task and higher efficiency due to belief sampling and symbolic planning integration. \par 2. Reduces computational costs by decreasing the number of search steps required for planning, improving efficiency by up to 26.8$\%$\\

\hline
DEPS \par \cite{wang2023describe} & Alfworld & Success Rate, Average Episode Length & 1. LLM generates PDDL goal from task description \par 2. Sampling world beliefs and using symbolic planning \par 3. Plan generator (BFS(f)) for generating plans & 1. LLM-DP achieves 96$\%$ success in Alfworld, outperforming the ReAct baseline (53$\%$) with fewer actions. \par 2.Reduces reliance on extensive LLM calls and minimizes human involvement by integrating automatic PDDL generation, thereby enhancing the efficiency of planning in dynamic environments.\\

\hline
PLaSma \par \cite{brahmanplasma} & COPLAN, VirtualHome & Human Evaluation: Coverage, Order, Overall Quality, Executability, Correctness & 1. Symbolic procedural knowledge distillation \par 2. Multi-task and task-specific distillation \par 3. Verifier-guided step-wise beam search \par 4. Constrained and counterfactual replanning tasks & 1. PlaSma enhances smaller models with procedural knowledge, outperforming GPT-3 in tasks like goal-based planning and counterfactual replanning. Achieves up to 93$\%$ success in counterfactual replanning. \par 2.Demonstrates efficiency by enabling smaller models to perform complex planning tasks with reduced computational costs.\\

\hline
KnowNo \par \cite{ren2023robots} & Simulated and Real Robot Tasks & Success Rate, Help Reduction, Task Success with Human Help, Prediction Set Size & 1. Use of Conformal Prediction (CP) for uncertainty calibration \par 2. Selection of multiple candidate actions \par 3. Minimizes human intervention based on prediction set size & 1. KNOWNO achieves 85$\%$ task success while reducing human help by 10–24$\%$. Guarantees task success with a calibrated confidence level using CP. \par 2. Enhances efficiency and autonomy by reducing the need for human intervention, aligning uncertainty to ensure task success \\

\hline
AutoToS \par \cite{cao2024automating} & BlocksWorld, 24 Game, Sokoban, Mini Crosswords, PrOntoQA & Success Rate, Plan Accuracy, Number of Language Model Calls, Time to Solve & 1. Automates feedback process for search components 2. Uses soundness and completeness checks to refine the generated code \par 3.Incorporates both generic and domain-specific unit tests & 1. AutoToS achieves 100$\%$ accuracy across five domains with minimal human feedback. Significantly reduces the number of model calls and ensures soundness and completeness. \par 2.Improves efficiency by automating the feedback process, significantly reducing the number of LLM calls and human involvement.\\
\hline
\end{tabular}
}
\label{tab:efficiency}
\end{table*}

\end{document}